\documentclass[final]{cvpr}

\usepackage{times}
\usepackage{epsfig}
\usepackage{subfigure}
\usepackage{graphicx}
\usepackage{amsmath}
\usepackage{amssymb}
\usepackage{multicol}
\usepackage{array, multirow}
\usepackage{rotating}
\usepackage{float}
\usepackage[bottom]{footmisc}
\usepackage{siunitx}
\usepackage{listings}

\usepackage{caption}

\usepackage[pagebackref=true,breaklinks=true,colorlinks,bookmarks=false]{hyperref}

\def\cvprPaperID{7781} 
\def\confYear{CVPR 2021}

\newcommand{\degree}{\ensuremath{^\circ}}

\begin{document}

\title{Large-scale Localization Datasets in Crowded Indoor Spaces}

\author{Donghwan Lee$^{1,}$\thanks{These authors contributed equally.} , 
        Soohyun Ryu$^{1,}$\footnotemark[1] ,
        Suyong Yeon$^{1,}$\footnotemark[1] ,
        Yonghan Lee$^{1,}$\footnotemark[1] ,
        Deokhwa Kim$^{1}$,
        Cheolho Han$^{1}$,
        \and
        Yohann Cabon$^{2}$,
        Philippe Weinzaepfel$^{2}$,
        Nicolas Guérin$^{2}$,
        Gabriela Csurka$^{2}$,
        and Martin Humenberger$^{2}$ \\
        \\
        $^{1}$NAVER LABS, $^{2}$NAVER LABS Europe\\
        {\tt\small $^{1,2}$\{donghwan.lee, soohyun.ryu, suyong.yeon, yh.l, deokhwa.kim, cheolho.han, yohann.cabon,} \\
        {\tt\small philippe.weinzaepfel, nicolas.guerin, gabriela.csurka, martin.humenberger\}@naverlabs.com}
}

\maketitle
\thispagestyle{empty}

\begin{abstract}
   Estimating the precise location of a camera using visual localization enables interesting applications such as augmented reality or robot navigation. This is particularly useful in indoor environments where other localization technologies, such as GNSS, fail. Indoor spaces impose interesting challenges on visual localization algorithms: occlusions due to people, textureless surfaces, large viewpoint changes, low light, repetitive textures, etc. Existing indoor datasets are either comparably small or do only cover a subset of the mentioned challenges. In this paper, we introduce 5 new indoor datasets for visual localization in challenging real-world environments. They were captured in a large shopping mall and a large metro station in Seoul, South Korea, using a dedicated mapping platform consisting of 10 cameras and 2 laser scanners. In order to obtain accurate ground truth camera poses, we developed a robust LiDAR SLAM which provides initial poses that are then refined using a novel structure-from-motion based optimization. We present a benchmark of modern visual localization algorithms on these challenging datasets showing superior performance of structure-based methods using robust image features. The datasets are available at: \url{https://naverlabs.com/datasets}    
\end{abstract}

\begin{figure}[h!]
    \centering
    \includegraphics[width=0.45\textwidth]{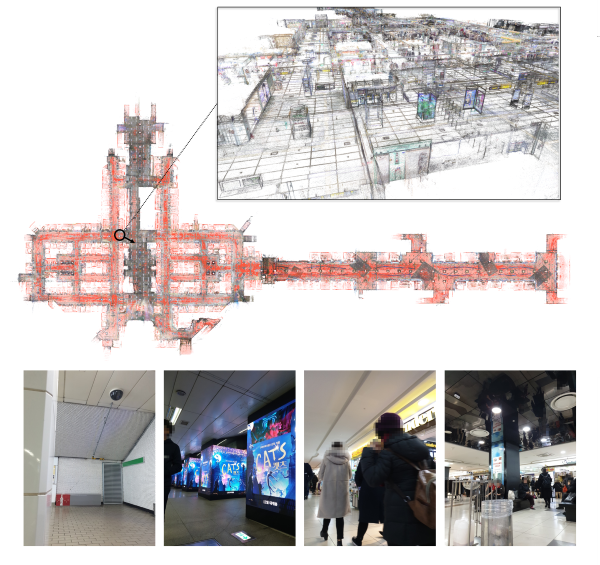} \\[-0.3cm]
    \caption{Illustration of the proposed NAVER LABS indoor visual localization datasets. \textit{Top:} Point clouds from LiDAR SLAM (red) and dense reconstruction~\cite{SchonbergerCVPR16StructureFromMotionRevisited} (color). \textit{Bottom:} Images showing different challenges; from left to right: textureless areas, changing appearance, crowdedness/occlusions, dynamic environment.}
    \vspace{-0.7cm}
    \label{fig:map_intro}
\end{figure}

\section{Introduction}
\label{sec:intro}
Visual localization estimates the 6 degrees of freedom (DoF) camera pose of a given query image. It is an important task in robotics, autonomous driving, and augmented reality (AR) applications. Modern visual localization methods that utilize both, deep learning and multi-view geometry, show promising results in outdoor environments, even in the presence of dramatic changes in scene appearance~\cite{ChenCVPR11CityScaleLandmarkIdentification,LiECCV12WorldwidePoseEst,Torii2019TPAMI,SattlerCVPR18Benchmarking6DoFOutdoorLoc}, and interesting real-world AR applications, \eg navigation, have been developed~\cite{GoogleVPS}. When considering challenges for visual localization, indoor environments differ from outdoor scenes in important aspects. For example, while the lighting conditions in indoor spaces tend to be more constant, indoor spaces are typically narrower than outdoor environments, thus, small camera movements can cause large changes in viewpoint. Dynamic objects, \eg, people, trolleys, flat screens, textureless areas, and repetitive patterns also impose challenges for visual localization and tend to occur more often indoors. 
High-accuracy indoor visual localization is crucial, especially for applications such as robot navigation, where a 10cm error in position could already cause a robot falling down a stair case. Even if in practice such extreme cases are avoided with proper redundancy in the on-board sensor suite~\cite{Ghani17IJARS,Baum18DSCC}, we want to push the capabilities of visual localization to its limits.
Large-scale indoor datasets with accurate ground truth poses are thus required in order to develop and benchmark high performing (\eg error below 10cm and 1\degree) visual localization algorithms.

Numerous datasets have been proposed for
visual localization~\cite{KendallICCV15PoseNetCameraRelocalization, LiECCV10LocationRecPriorFeatureMatching, SattlerBMVC12ImRetLocalizationRevisited, SattlerCVPR18Benchmarking6DoFOutdoorLoc,SattlerBMVC12ImRetLocalizationRevisited, LiECCV12WorldwidePoseEst, IrscharaCVPR09FromSFMLocationRecognition, IrscharaCVPR09FromSFMLocationRecognition, ChenCVPR11CityScaleLandmarkIdentification,Torii2019TPAMI, Maddern2017IJRR, Badino_IV11, Glocker2013ISMAR, Taira2018CVPR} and Table~\ref{tab:dataset_compare} provides an overview.
In outdoor environments, structure-from-motion (SFM) is employed as an effective tool to generate the ground truth poses. Unfortunately, due to the aforementioned challenges, large-scale SFM is more difficult to apply for indoor spaces~\cite{Walch16ICCV,Sun2017CVPR}. Thus, indoor datasets often rely on additional depth sensors such as RGB-D cameras~\cite{Glocker2013ISMAR,Taira2019TPAMI} and laser scanners (LiDAR)~\cite{Sun2017CVPR}. The depth range provided by RGB-D cameras is often limited to about 10m~\cite{IntelRealsenseD435}. As a result, most of the existing datasets constructed using RGB-D cameras capture relatively small indoor spaces~\cite{Armeni20163d,Dai2017CVPR,Chang20173DV,Glocker2013ISMAR,Taira2018CVPR} and LiDARs~\cite{Sun2017CVPR} or dedicated visual 3D mapping devices~\cite{Taira2019TPAMI,Chang20173DV} have been used in order to create larger indoor datasets.
Such devices generate panoramic images or colored point clouds that capture the structure of indoor spaces with high spacial resolution~\cite{ZebDiscovery,NavVisM6}. 
In addition to high precision 3D scans, a good visual localization dataset may also consist of many images covering a large variety of views and situations.
Thus, additional training and realistic query images often have to be recorded separately and they need to be registered within the generated point clouds in order to compute their ground truth poses.
One solution for this, as already done for existing indoor visual localization datasets~\cite{Sun2017CVPR}, is utilizing semi-automatic annotation algorithms with a significant amount of manual workload.
As a consequence, the area covered by existing datasets (see Table~\ref{tab:dataset_compare}) is sparsely sampled by the provided images~\cite{Glocker2013ISMAR,Taira2018CVPR} and this causes difficulties when, \eg, training deep neural networks for visual localization~\cite{SattlerCVPR19UnderstandingLimitationsPoseRegression}.
We conclude that datasets that are large-scale, accurate, and densely sampled are still missing.

This motivated us to contribute another step towards closing this gap.
In this paper, we introduce five new localization datasets acquired in challenging real-world indoor environments. The datasets were captured in a large shopping mall and a large metro station in Seoul, South Korea and include images and point clouds obtained from a dedicated mapping platform comprising ten cameras and two 3D laser scanners. Compared to existing indoor datasets, the proposed datasets provide dense image sampling and cover many challenges of indoor visual localization (see Figure~\ref{fig:map_intro}). In addition, the database and the query sets were captured with time intervals of up to 128 days, thus, the datasets contain various changes over time. 
Since estimating precise ground truth (6DoF) camera poses for such large indoor spaces using SFM, to the best of our knowledge, was not possible so far (\eg \cite{Walch16ICCV,Sun2017CVPR} report failure cases), we developed a novel, automated pipeline that utilizes both, LiDAR SLAM and SFM algorithms to solve this problem. 
The proposed pipeline applies LiDAR-based pose-graph SLAM to estimate the trajectories of the mapping platform. 
The LiDAR SLAM poses are spline-smoothed and used as a prior information for the subsequent SFM process that obtains the camera poses jointly for multiple trajectories within the same space. 
As shown in Figure~\ref{fig:map_intro}, the 3D models from both LiDAR SLAM and SFM are well aligned. 
The datasets are referred to as \emph{NAVER LABS localization datasets} and are available at: \\
\url{https://naverlabs.com/datasets}

Our primary contributions are summarized as follows.
\textit{(i)} We introduce five new large-scale datasets for challenging real-world indoor environments (Section~\ref{sec:datasets}). 
The proposed datasets provide dense image sampling with ground truth poses as well as accurate 3D models. 
\textit{(ii)} We propose a novel, fully-automated pipeline based on LiDAR SLAM and SFM to obtain ground truth poses (Section~\ref{sec:processing}). 
\textit{(iii)} We present detailed evaluations of modern visual localization methods of relevant algorithmic groups showing their limits on these challenging datasets (Section~\ref{sec:benchmark}).
We hope to open new interesting research directions to further improve visual localization.

\begin{table*}[]
    \centering
    \scriptsize{
    \begin{tabular}{clcccccccc}
        \hline
         \multicolumn{2}{c}{\textbf{Datasets}} & \textbf{Environments} &  \multicolumn{2}{c}{\textbf{Images}} &  \textbf{Space}  & \multicolumn{2}{c}{\textbf{3D SFM Model}} & \multicolumn{2}{c}{\textbf{Condition}} \\
         & & &  \textbf{DB} & \textbf{Query} & \textbf{($m^2$)} & \textbf{Points} & \textbf{Observations} & \textbf{Interval} & \textbf{Crowdedness} \\
         \hline
         \multirow{10}*{\begin{sideways} Outdoor \end{sideways}}  
         & Cambridge~\cite{KendallICCV15PoseNetCameraRelocalization} & 5 smaller scenes &  6,848 & 4,081 & & 1.89M & 17.68M & - & -\\
         & Dubrovnik~\cite{LiECCV10LocationRecPriorFeatureMatching} & large city area  & 6,044 & 800 & & 1.89M & 9.61M & - & -\\
         & Aachen Day-Night~\cite{SattlerCVPR18Benchmarking6DoFOutdoorLoc,SattlerBMVC12ImRetLocalizationRevisited,Zhang2020ARXIV} & medium city area & 4,328 & 922 & & 1.65M & 10.55M & - & -\\
         & Landmarks~\cite{LiECCV12WorldwidePoseEst} & 1k landmarks & 204,626 & 10,000  &  & 38.19M & 177.82M & - & - \\
         & Rome~\cite{LiECCV10LocationRecPriorFeatureMatching} & 69 landmarks & 15,179 & 1000 &  & 4.07M & 21.52M & - & - \\
         & Vienna~\cite{IrscharaCVPR09FromSFMLocationRecognition} &  3 city zones  &  1,324 & 266 &  & 1.12M & 4.85M & - & - \\
         & San Francisco~\cite{ChenCVPR11CityScaleLandmarkIdentification,LiECCV12WorldwidePoseEst,Torii2019TPAMI} &  very large city area  &  1M & 803 & & 75M & 300M & - & - \\
         & RobotCar Seasons~\cite{SattlerCVPR18Benchmarking6DoFOutdoorLoc,Maddern2017IJRR} & 49 urban zones  &  20,862 & 11,934 & & 6.77M & 36.15M & - & - \\
        & CMU Seasons~\cite{SattlerCVPR18Benchmarking6DoFOutdoorLoc,Badino_IV11}  & 17 sub-urban zones &  7,159 & 75,335  & & 1.61M & 6.50M & - & - \\
        \hline
        \multirow{6}*{\begin{sideways} Indoor \end{sideways}}
        & Stanford~\cite{Armeni20163d} & 270 small rooms &  \multicolumn{2}{c}{71,909} & 6,020 & \multicolumn{2}{c}{laser scanner} & - &  -\\
        & ScanNet~\cite{Dai2017CVPR} & 707 small rooms &  \multicolumn{2}{c}{2,492,518} & 34,453 & \multicolumn{2}{c}{RGB-D camera} & - & -\\
        & Matterport3D~\cite{Chang20173DV} & 2,056 small rooms & \multicolumn{2}{c}{194,000} & 46,561 & \multicolumn{2}{c}{RGB-D camera} & - & - \\
        & 7-scenes~\cite{Glocker2013ISMAR} & 7 small rooms & 26,000 & 17,000 & - & \multicolumn{2}{c}{RGB-D camera} & - & $<$ 0.1\% \\
        & TUM Indoor~\cite{Huitl12ICIP} & Univ.bldg. (7 floors) & 41,888 & 7,086 & 16,341 & \multicolumn{2}{c}{laser scanner} & 26 days & 2.1\% \\
        & TUM-LSI~\cite{Walch16ICCV} & Univ. bldg. (5 floors) & 875 & 220 & 5,575 & \multicolumn{2}{c}{laser scanner} & - & - \\
        & InLoc~\cite{Taira2018CVPR} & Univ. bldg. (5 floors) & 9,972 & 356 & 10,370 & \multicolumn{2}{c}{laser scanner / RGB-D} & months & $<$ 0.1\% \\
        & Baidu~\cite{Sun2017CVPR}  & mall & 682 & 2,296 & 9,179 & \multicolumn{2}{c}{laser scanner} & - & 3.75\% \\
        \hline
        \multirow{5}*{\begin{sideways} \textbf{NL (Ours)} \end{sideways}}
         & Dept. B1 & mall & 22,726 & 10,757 & 8,513 & 1.7M & 17.1M & 128 days & 6.87\% \\
         & Dept. 1F & mall & 21,600 & 5,323 & 10,046 & 1.5M & 10.6M & 128 days & 5.47\% \\
        & Dept. 4F & mall  & 12,421 & 7,515 & 8,348 & 1.3M & 8.7M & same day &  3.45\% \\
         & Metro St. B1 & mall \& turnstiles  & 23,795 & 17,638 & 20,879 & 3.1M & 13.6M & 17 days &  12.9\% \\
         & Metro St. B2 & platform & 6,768 & 8,240 & 5,250 & 1.3M & 4.8M & same day & 7.73\% \\
        \hline
    \end{tabular}
    }
    
    \vspace{-0.2cm}
    
    \caption{Visual localization datasets. \textbf{Interval} represents the time interval between the capture of the database and the query sets. \textbf{Crowdedness} is the percentage of pixels belonging to humans. We compute this with an off-the-shelf semantic segmentation network~\cite{Deeplabv3plus2018} followed by manual inspection. NL stands for NAVER LABS.}
    \vspace{-0.3cm}
    
    \label{tab:dataset_compare}
\end{table*}

\section{The NAVER LABS localization datasets}
\label{sec:datasets}
This section first describes the characteristics of the proposed datasets as well as the considered indoor spaces, it then introduces the mapping platform we used to acquire the data, and finally, it relates our datasets with existing work.

\subsection{Dataset description} 
Our datasets consist of multiple sequences captured over four months in five different floors in a department store and a metro station (see Table~\ref{tab:dataset_compare}).
All places were publicly accessible while scanning, thus many moving objects such as people, trolleys, luggage, as well as flat screens and store fronts were captured. Each floor is different in terms of size, statistics, and characteristics: The captured surface size varies from 5,250$m^2$ to 20,879$m^2$. The entire dataset contains over 130k images, each individual floor contains between 14k and 41k images. The 3D models, reconstructed with the proposed SFM pipeline, consist of 1.3M to 3.0M 3D points, triangulated from 4.8M to 22.7M local features (see Table~\ref{tab:dataset_compare}).  

The \textit{Dept.}~datasets were captured in one of the largest shopping malls in South Korea. It consists of three different floors: \textit{Dept.~B1} contains areas with restaurants, supermarkets, and cafes, \textit{Dept.~1F} and \textit{4F} contain cosmetic, jewelry, and clothing shops. \textit{Dept.~1F} and \textit{4F} were captured in good lighting conditions but include lots of reflective surfaces. \textit{Dept.~B1} was captured under low-light conditions. During data acquisition, \textit{Dept.~4F} was under construction, thus, a large number of images containing texture-less temporary walls are included. Rapid rotations of the mapping platform often caused motion blur in the images. 

\textit{Metro~St.}~was collected in one of the most crowded metro stations in Seoul and is intended for evaluating robustness of visual localization algorithms in scenes with many moving objects. The images in this dataset show the largest proportion of human pixels among the existing indoor localization datasets, including ours. Furthermore, this dataset is the largest in size and includes a variety of scenes with shops, turnstiles, and stairs. \textit{Metro~St.~B2} was captured at the platforms of the metro station. The two platforms in this dataset were built with highly similar designs and internal structures as shown in Figure~\ref{fig:Gangnam}. This dataset is intended for benchmarking visual localization algorithms in repetitive and symmetric scenes. In addition, digital signage and platform screen doors, which change appearance over time, introduce new challenges. 

\subsection{Dataset format}
To easily evaluate an image-based localization technique, each dataset is provided in the \emph{kapture}\footnote{\url{https://github.com/naver/kapture}} format~\cite{humenberger2020robust} which supports timestamps, shared camera parameters, multi-camera rigs, reconstruction data, and other sensor data. The kapture toolbox enables to convert data into other formats such as OpenMVG~\cite{MoulonOpenMVG} and COLMAP~\cite{schoenberger2016sfm}. 

\begin{figure}
    \centering
    \includegraphics[width=0.8\linewidth]{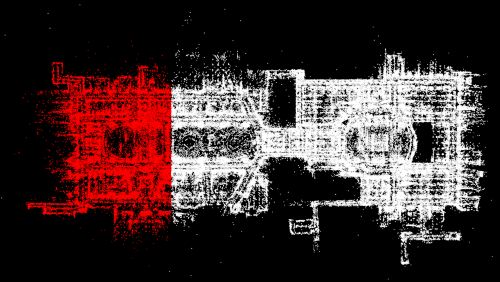} 
    \caption{Example of validation (red) and test (white) splits for one of our datasets (Dept.~4F).}
    \vspace{-0.05cm}
    \label{fig:testvalsplit}
\end{figure}

As can be seen in Table~\ref{tab:dataset_compare}, our datasets consist of database images for mapping, and query images for testing. In addition, we split the query images into test and validation sets that represent different areas with little overlap (see Figure~\ref{fig:testvalsplit}). The ground truth poses of the validation set are released publicly, the poses of the test set are retained to build a visual localization benchmark.

\subsection{Mapping platform} \label{sub:mapping_platform}

\begin{table*}[]
    \begin{center}
    \scriptsize
    \begin{tabular}{cccc}
        \hline\hline  
        \textbf{Type} & \textbf{Amount} & \textbf{Model} & \textbf{Specification}  \\
        \hline
        3D LiDAR & 2 & Velodyne VLP-16 & FoV: 360\si{\degree}$\times$30\si{\degree}, angular resolution: 0.2\si{\degree}$\times$2.0\si{\degree}, rotation rate: 10Hz,   \\
        camera & 6 & Basler acA2500-20gc & FoV: 79.4\si{\degree}$\times$63.0\si{\degree}, image resolution: 2592$\times$2048, frame rate: 2.5Hz, global shutter \\
        smartphone (as camera) & 4 & Galaxy S9 & FoV: 77.0\si{\degree}(D), image resolution: 2160$\times$2880, frame rate: 1Hz, rolling shutter \\
        \begin{tabular}{c}
        magnetic rotary position sensor \\ (as wheel encoder) 
        \end{tabular} & 2 & ams AS5047 & resolution: 1024 pulses/rotation\\
        \hline
    \end{tabular}
    
    \vspace{-0.2cm}
    
    \caption{Sensors of the platform.}
    \label{tab:sensors}
    
    \vspace{-0.5cm}
    
    \end{center}
\end{table*}

\begin{figure}
    \centering
    \subfigure[]{\includegraphics[width=0.15\textwidth]{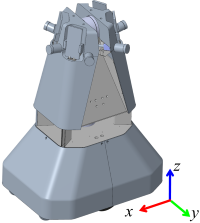}\label{fig:subfig2a}}
    \hspace{.1cm}
    \subfigure[]{\includegraphics[width=0.15\textwidth]{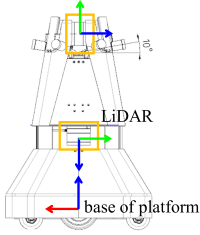}\label{fig:subfig2b}}
    \subfigure[]{\includegraphics[width=0.15\textwidth]{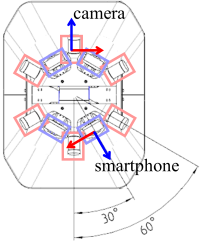}\label{fig:subfig2c}}
    \vspace{-0.4cm}
    \caption{The mapping platform: \subref{fig:subfig2a} appearance, \subref{fig:subfig2b} side view \subref{fig:subfig2c}, top view.}
    \label{fig:platform}
    
    \vspace{-0.3cm}
    
\end{figure}

We collected data by using the dedicated mapping platform shown in Figure~\ref{fig:platform}. The mapping platform is equipped with differential drive consisting of two wheels and four casters. Each wheel has an encoder to obtain odometry data. The center of the virtual axle connecting two wheels is considered as the origin of the platform, the \emph{x}-axis defines the forward direction, and the \emph{z}-axis defines the opposite of gravity.

The platform is equipped with two LiDARs with complementary roles: One LiDAR is positioned horizontally at the center of the platform because horizontal orientation of a LiDAR sensor maximizes the information required to estimate poses in indoor spaces. The other LiDAR is positioned perpendicularly at the top of the platform to construct the dense 3D point clouds via push broom scanning. 

Furthermore, the mapping platform is equipped with two different types of cameras; 6 industrial cameras (Basler acA2500-20gc) and 4 smartphone (Galaxy S9) cameras. The six industrial cameras are mounted at intervals of 60 degrees. They have a horizontal field of view (FoV) of more than 70 degrees, thus, a set of images can horizontally cover 360 degrees. Each camera is facing 10 degrees upwards in order to capture not too much ground surface. The four smartphones are mounted slightly above the industrial cameras to better reflect real-world applications where the capture device at test time will be different from the mapping platform. The platform and the sensors are synchronized using a network time protocol (NTP) server. Note that illumination can change easily with camera movement in indoor spaces, thus, we set the exposure time to automatic for all cameras. Detailed specifications of each sensor are given in Table~\ref{tab:sensors}. 

For extrinsic calibration between the two LiDARs, we estimate the relative pose using the iterative closest point (ICP) algorithm~\cite{besl1992pami, Segal2009RSS} using geometric features in their shared field of view. For calibration between the base and the cameras, we employ an online self-calibration via SFM rather than an offline method. This is motivate by the fact that, on the mapping platform, the smartphone cameras have irregular time delays when compared to the industrial cameras caused by automatic exposure time control and SW synchronization of Android OS. As a result, an online calibration method, which can deal with irregular time offsets, is better suited. Additional details and results are described in Section~\ref{sub:SFM}.

\subsection{Related datasets}
\label{sec:datasets}

To complete the description of the NAVER LABS datasets, in this section we discuss a list of related work.
Existing datasets often used for visual localization can be divided in indoor and outdoor (see Table~\ref{tab:dataset_compare}). 
While both groups share the fundamental challenges of visual localization (Section~\ref{sec:intro}), outdoor datasets often contain images including environmental changes, \textit{e.g.}, seasons, weather, illumination, and indoor datasets often include images that represent occlusions, dynamic objects, and repetitive structures. 
Popular indoor datasets such as Stanford~\cite{Armeni20163d}, ScanNet~\cite{Dai2017CVPR}, and Matterport3D~\cite{Chang20173DV} target semantic understanding rather than visual localization, thus, they do not provide database and query splits. 
However, even indoor datasets~\cite{Glocker2013ISMAR,Taira2019TPAMI,Sun2017CVPR} that are more focused on localization tasks were captured in a comparably small indoor spaces which makes them less relevant for real-world applications. 
Another important aspect of indoor localization is crowdedness in the sense of image regions covered by people.
This is interesting because it effects both, training and testing (\ie mapping and localization).
As can be seen in Table~\ref{tab:dataset_compare}, crowdedness has not been considered so far.
In summary, large-scale datasets for visual localization are more often available outdoors, in addition, indoor datasets rarely contain people as inherent part of the environments they cover. 
The proposed datasets target on closing these gaps. 
They are captured in highly dynamic and crowded indoor spaces and contain a large variety of symmetric, non-Lambertian, and texture-less areas. 
Furthermore, since these datasets were captured during business hours, they better reflect real-world applications.

\section{Ground truth generation}
\label{sec:processing}

The ground truth poses for all images are generated using our LiDAR SLAM (Section~\ref{sub:slam}) and SFM algorithms (Section~\ref{sub:SFM}). First, LiDAR SLAM produces poses of the mapping platform for each sequence. Then it merges sequences taken in the same space such that the sequences have a unified coordinate system. Note that the point clouds and camera poses can then be computed based on the platform poses and calibration parameters. Finally, SFM is used to refine the camera poses. This process allows us to obtain extrinsic camera calibration of the mapping platform without the need of a manual calibration process.

\subsection{LiDAR SLAM}
\label{sub:slam}

In order to estimate the platform poses, we developed a LiDAR SLAM pipeline based on pose-graph optimization~\cite{LuAuro1997}.
In the pose-graph, a node represents a platform pose and an edge between two nodes represents a relative pose constraint associated with the two platform poses. 

\begin{figure}
    \centering
    \includegraphics[width=0.45\textwidth]{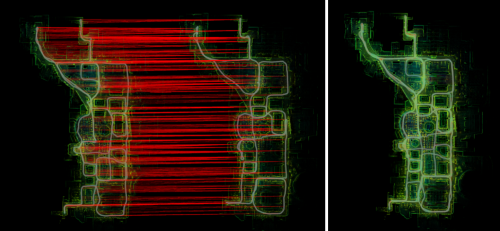} \\[-0.3cm]
    \caption{Pose-graph for different sequences acquired at \textit{Dept.~B1} before (left) and after (right) merging and optimization. The thick white line represents platform poses, the green dots represent the accumulated point cloud according to the poses, and the red line represents a connection between spatially adjacent nodes in different sequences.}
    \label{fig:pose-graph}
    
    \vspace{-0.3cm}
    
\end{figure}

\noindent \textbf{Undistorted point cloud generation.}
To increase the amount of point cloud data used in one node, we merge point clouds collected from the two LiDARs. We transform point clouds represented in different LiDAR coordinate systems into a platform coordinate system (using the extrinsic offline calibration of the LiDARs) and concatenate them. In addition, in order to collect point clouds efficiently and evenly, we acquire point clouds by moving the platform at speeds of up to 1.0 $m/s$. The scanning LiDAR characteristics, which are similar to those of a rolling shutter camera, cause distortion of point clouds while being in motion. We employed wheel odometry to compensate for that. After calculating the transforms between platform poses at different times from wheel odometry, they are applied to the points to execute the compensation. Here, the position of a point in a point cloud in the platform coordinate system is calculated as
\begin{equation}
{}^{B_t}\tilde{\mathbf{p}}=\tilde{\mathbf{R}}_{B_tB_{t+\Delta t}}(\mathbf{R}_{BL}{}^{L_{t+\Delta t}}\tilde{\mathbf{p}}+\mathbf{t}_{BL})+\tilde{\mathbf{t}}_{B_tB_{t+\Delta t}},
\label{eq:tf}
\end{equation}
where $B_t$ and $L_t$ define the base of the mapping platform and the LiDAR at time $t$, respectively, and $[\tilde{\mathbf{R}}_{AB}|\tilde{\mathbf{t}}_{AB}]$ is an odometry measurement by the wheel encoder from $A$ to $B$. Thus, (\ref{eq:tf}) transforms the LiDAR point ${}^{L_{t+\Delta t}}\tilde{\mathbf{p}}$ at time $t+\Delta t$, to the platform point ${}^{B_t}\tilde{\mathbf{p}}$.

\noindent \textbf{Pose-graph SLAM.}
Since the wheel odometry is comparably inaccurate, we set the edges of the graph as relative poses estimated from ICP using undistorted sequential point clouds.
However, it is difficult to compensate for accumulated pose errors using the information between sequential nodes only. To overcome this, additional edges between spatially-adjacent but time-distant nodes are calculated using ICP and added to the pose-graph.
We acquire these edges by following a coarse-to-fine approach. Using a distance threshold, we first select candidates that we quickly verify with a rough ICP. Then we refine the verified candidates using a precise ICP. These edges act as additional bridges to increase robustness and enable loop closing. The platform poses are then estimated by optimizing this graph.

\noindent \textbf{Graph merging.}
Platform poses estimated by the above procedures are represented in an independent coordinate system for each sequence. To unify the different coordinate systems, we perform a merging process between graphs from each sequence (Figure~\ref{fig:pose-graph}). As we start to scan a new sequence roughly at the same starting position like the other ones, the graphs are already coarsely aligned and loop closing (via ICP) can be performed throughout all the graphs (we needed to manually correct only a few initial poses). By optimizing the graph, we obtain the platform poses represented in a unified coordinate system.

\subsection{Structure-from-motion} \label{sub:SFM}
In this section, we describe the SFM pipeline used to refine the initial image poses from LiDAR SLAM and to reconstruct the 3D feature map aligned to the LiDAR point cloud. The proposed bundle adjustment utilizes the prior LiDAR SLAM poses and employs incremental mapping of the COLMAP library~\cite{SchonbergerCVPR16StructureFromMotionRevisited} to perform accurate and efficient large-scale indoor SFM. The intrinsic and rotational extrinsic parameters were also auto-calibrated for each dataset during the SFM process.

\noindent \textbf{Local feature matching.}
We extracted R2D2~\cite{RevaudNIPS19R2D2ReliableRepeatableDetectorsDescriptors} features for each image. 
Then, we matched the features efficiently using only the spatially close image pairs.
More precisely, we use the initial image poses from the LiDAR SLAM process to effectively select most of the co-visible image pairs by thresholding the distance and the viewing angle difference between two images. We then filter the initial feature matches using cross-validation and reject outliers using geometric verification.

\begin{figure}[t!]
    \centering
    \includegraphics[width=0.5\textwidth]{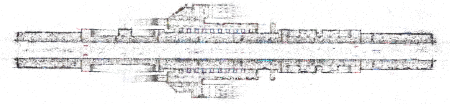} \\[-0.3cm]
    \caption{3D reconstruction of Metro St. B2.}
    \vspace{-0.3cm}
    \label{fig:Gangnam}
\end{figure}

\noindent \textbf{Map optimization.}
Although traditional SFM libraries, such as COLMAP~\cite{SchonbergerCVPR16StructureFromMotionRevisited} and Visual SFM~\cite{Wu-2011-Visualsfm}, could generally be applied to construct and optimize a 3D point cloud map, we found that their standard bundle adjustment implementation was not suited for our purpose, \ie, refinement of LiDAR SLAM poses. To perform SFM precisely and efficiently in such large indoor spaces, it is crucial to embed precise prior pose information and rigid sensor configuration constraints. In the existing SFM optimizers, however, this is typically not possible if images or prior poses are not time-synchronized. Thus, we cannot apply them because our datasets contain various types of sensors, \ie, industrial cameras, smartphone cameras, and LiDARs, which are hard to synchronize.

For this reason, we developed a novel bundle adjustment algorithm which can optimize image poses, feature points, and calibration parameters, jointly. This is done by utilizing poses from the continuous trajectories of the platform that are estimated using LiDAR SLAM, as priors. To account for timestamp difference between sensors, we represented each platform trajectory as a continuous cubic spline in SE(3)~\cite{mueggler2018ctvio}. With the platform trajectory spline $\bold{T}_{WB}(t)$ and the extrinsic pose for camera $C_i$, $\bold{T}_{BC_i}$, the pose error can be written as
\begin{equation}
\bold{e}^{spline}_{i} = \bold{T}_{WC_i} \ominus (\bold{T}_{WB}(t_{i}) \; \bold{T}_{BC_i}).
\end{equation}
Here, $\ominus$ denotes generalized minus defined in SE(3) manifold. $\bold{T}_{WC_i}$ is the image pose and $t_i$ the corresponding timestamp. This continuous representation of the platform trajectory enables efficient and elegant usage of asynchronous prior poses coming from LiDAR SLAM. More importantly, rigid configuration constraints between the cameras and the platform can be embedded even though the images of two different types of cameras were not captured time-synchronously. The proposed pose error via spline can be optimized with feature reprojection error 
\begin{equation}
\bold{e}^{proj}_{ij} = \Tilde{\bold{z}}_{ij} - \pi_{C_i}(\bold{T}_{C_iW} \; {}^W\bold{p}_j)
\end{equation}
to minimize the joint cost function
\begin{equation}
\mathcal{L} = \sum_{i\in\mathcal{I}} \left[ \rho(\|\bold{e}^{proj}_{ij}\|^2) \; + \; \|\bold{e}^{spline}_{i}\|^2 \right],
\end{equation}
where $\rho$ denotes the Cauchy loss \cite{triggs1999bundle}. By this modified bundle adjustment method, image poses and triangulated feature points can be refined and auto-calibration of intrinsic and extrinsic parameters can be conducted.

\begin{table}
 \resizebox{\linewidth}{!}{
 \begin{tabular}{c c c c c}
 \hline
    \textbf{Dataset}  & \textbf{Images} & \textbf{Points} & \textbf{Avg. obs.} & \textbf{Rep. err.} \\
 \hline
    Dept. B1 & 33,483 & 1.7M & 510.8 & 1.3614 \\
    Dept. 1F & 26,923 & 1.5M & 394.5 & 1.3436 \\
    Dept. 4F & 19,936 & 1.3M & 436.5 & 1.4695 \\
 \hline
     Metro St. B1 & 41,433 & 3.1M & 329.2 & 1.0373 \\
     Metro St. B2 & 14,958 & 1.3M & 320.6 & 0.9900 \\
 \hline
 \end{tabular}
 }
 
 \vspace{-0.2cm}
 
 \caption{Mean reprojection error (Rep.~err.) and mean feature observation count in the images (Avg.~obs.) per dataset. }
 
 \vspace{-0.3cm}
 
 \label{tab:ImageStat}
\end{table}

\noindent \textbf{SFM pipeline.}
We generated the SFM models by iterating our bundle adjustment and COLMAP's incremental triangulation \cite{SchonbergerCVPR16StructureFromMotionRevisited}, and for each iteration we decreased the outlier filtering thresholds.
The initial reprojection errors using poses from LiDAR SLAM vary between 50 and 200 pixels. As shown in Table~\ref{tab:ImageStat}, our SFM achieves very low reprojection errors under 1.5 pixels.
Additionally, our SFM pipeline could accurately register images with low visual saliency or crowded people by incorporating prior LiDAR SLAM results. An example of our sparse and dense 3D reconstruction models are shown in Figure~\ref{fig:Gangnam} and Figure~\ref{fig:map_intro}, respectively.

\noindent \textbf{Auto-calibration.}
As mentioned above, the camera calibration parameters are automatically estimated while reconstructing the 3D model of each dataset via our SFM pipeline. To be more precise, only the intrinsic parameters and the rotational part of the extrinsic (camera to platform) parameters were auto-calibrated, the translation part of the extrinsic parameters is obtained from the CAD model. Motions of the mapping platform were much smaller (up to 1$m/s$), when compared to the scale of target indoor spaces (up to 50$m$). Experimentally, we found that the translation motion of the mapping platform showed a small impact on the reprojection loss term, which leads to unstable convergence of extrinsic parameters. Contrary, if we fix the translation part of extrinsic parameters, all camera parameters converge better.

\begin{figure}[t!]
    \centering
    \includegraphics[width=0.4\textwidth]{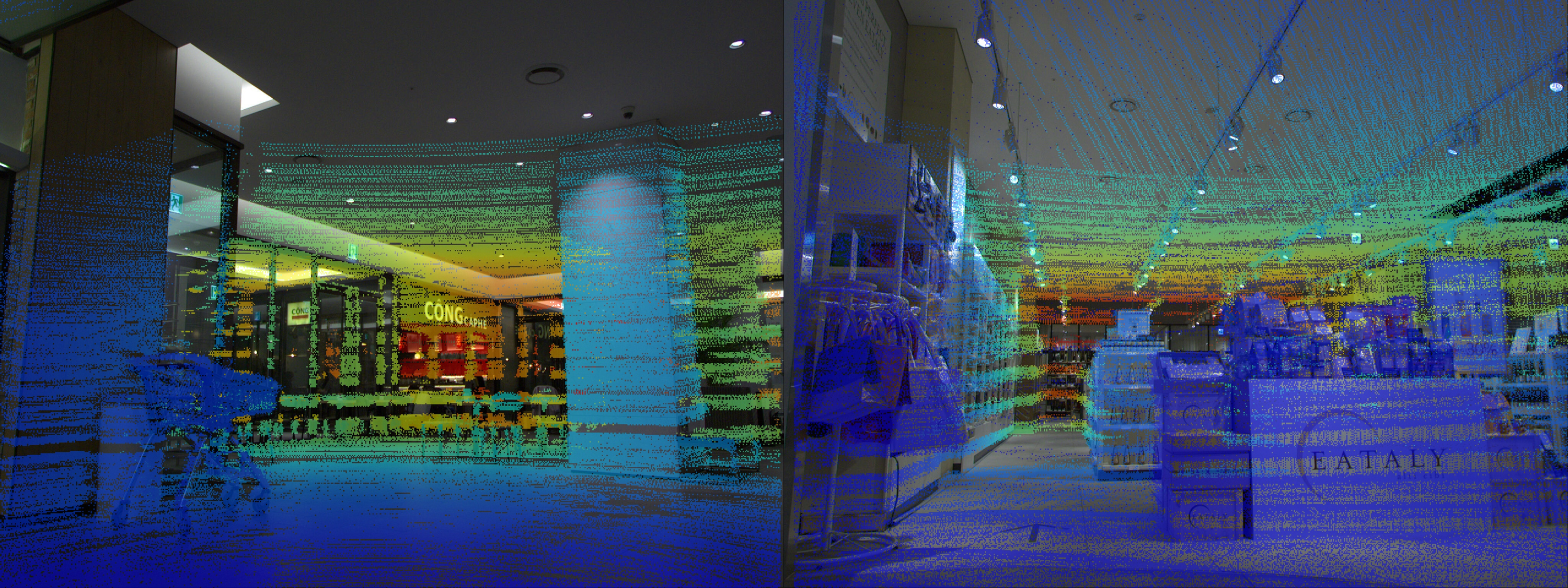} \\[-0.2cm]
    \caption{LiDAR scan data projected to images.}
    \label{fig:PcdDepth}
    
    \vspace{-0.3cm}
    
\end{figure}

\section{Visual localization benchmark}
\label{sec:benchmark}

\begin{table*}[ht!]
\resizebox{\linewidth}{!}{
\begin{tabular}{|l|rrr|rrr|rrr|rrr|rrr|}
\hline
\textbf{Test set - \emph{Galaxy} images} & \multicolumn{3}{c|}{ Dept. B1 } & \multicolumn{3}{c|}{ Dept. 1F } & \multicolumn{3}{c|}{ Dept. 4F } & \multicolumn{3}{c|}{ Metro St. B1 } & \multicolumn{3}{c|}{ Metro St. B2 }  \\
\textbf{Algorithm / Accuracy th.} & 0.1m,1\degree   & 0.25m,2\degree  & 1m,5\degree & 0.1m,1\degree   & 0.25m,2\degree  & 1m,5\degree & 0.1m,1\degree   & 0.25m,2\degree  & 1m,5\degree & 0.1m,1\degree   & 0.25m,2\degree  & 1m,5\degree & 0.1m,1\degree   & 0.25m,2\degree  & 1m,5\degree \\
\hline
\textit{structure-based methods} & &&&&&&&&&&&&&& \\
  ~~APGeM~\cite{RevaudICCV19LearningwithAPTrainingImgRetrievalListwiseLoss}+SIFT~\cite{Lowe04IJCV}           & 69.8 & 75.7 & 81.3 & 74.0 & 78.8 & 83.2 & 83.4 & 89.2 & 91.5 & 40.8 & 59.6 & 69.5 & 39.2 & 57.8 & 62.8  \\
  ~~DELG~\cite{CaoX20UnifyingDeepLocalGlobalFeatures}+SIFT~\cite{Lowe04IJCV}            & 71.2 & 77.6 & 84.0 & 79.1 & 83.7 & 88.3 & 81.2 & 87.0 & 89.2 & 40.7 & 60.3 & 70.2 & 42.1 & 60.3 & 65.6  \\
  ~~DenseVLAD~\cite{ToriiCVPR15PlaceRecognitionByViewSynthesis}+SIFT~\cite{Lowe04IJCV}       & 72.2 & 77.9 & 84.5 & 82.6 & 86.9 & 92.3 & 84.6 & 89.8 & 92.3 & 40.8 & 60.4 & 69.7 & 41.8 & 59.2 & 64.2  \\
  ~~NetVLAD~\cite{ArandjelovicCVPR16NetVLADPlaceRecognition}+SIFT~\cite{Lowe04IJCV}         & 72.0 & 78.2 & 84.8 & 82.2 & 86.2 & 90.3 & 86.5 & 92.0 & \underline{94.9} & 38.6 & 55.8 & 66.3 & 41.9 & 59.9 & 65.3  \\
  ~~SIFT~\cite{Lowe04IJCV}+vocab.~tree (COLMAP~\cite{schoenberger2016sfm})           & 67.6 & 75.0 & 80.7 & 80.8 & 84.9 & 88.9 & 68.7 & 73.9 & 76.2 & 38.0 & 57.2 & 67.0 & 39.0 & 56.7 & 62.5  \\
  ~~APGeM~\cite{RevaudICCV19LearningwithAPTrainingImgRetrievalListwiseLoss}+D2-Net~\cite{DusmanuCVPR19D2NetDeepLocalFeatures}         & 73.7 & 79.3 & 87.2 & 78.0 & 82.8 & 88.0 & 84.2 & 89.8 & 92.0 & 44.2 & 64.8 & 75.7 & 41.3 & 60.1 & 65.0  \\
  ~~DELG~\cite{CaoX20UnifyingDeepLocalGlobalFeatures}+D2-Net~\cite{DusmanuCVPR19D2NetDeepLocalFeatures}          & 75.3 & 81.7 & \underline{91.6} & 84.5 & 89.2 & 94.3 & 81.8 & 87.2 & 90.0 & 44.3 & 65.4 & 76.0 & 40.8 & 60.5 & 65.9  \\
  ~~DenseVLAD~\cite{ToriiCVPR15PlaceRecognitionByViewSynthesis}+D2-Net~\cite{DusmanuCVPR19D2NetDeepLocalFeatures}     & 75.7 & 81.5 & 90.5 & 85.2 & 89.7 & \underline{95.9} & 85.0 & 90.5 & 92.9 & 44.4 & 64.8 & 74.6 & 42.3 & 60.5 & 65.1  \\
  ~~NetVLAD~\cite{ArandjelovicCVPR16NetVLADPlaceRecognition}+D2-Net~\cite{DusmanuCVPR19D2NetDeepLocalFeatures}        & 75.5 & \underline{82.3} & 91.4 & 84.5 & 88.9 & 94.2 & \underline{87.0} & \underline{92.4} & 94.9 & 42.9 & 61.9 & 71.3 & 41.7 & 61.5 & 66.3  \\
  ~~APGeM~\cite{RevaudICCV19LearningwithAPTrainingImgRetrievalListwiseLoss}+R2D2~\cite{RevaudNIPS19R2D2ReliableRepeatableDetectorsDescriptors}           & 75.2 & 80.3 & 87.6 & 80.6 & 84.3 & 89.4 & 85.3 & 91.0 & 93.1 & \underline{46.5} & \underline{66.8} & 76.7 & 43.2 & 62.0 & 66.3  \\
  ~~DELG~\cite{CaoX20UnifyingDeepLocalGlobalFeatures}+R2D2~\cite{RevaudNIPS19R2D2ReliableRepeatableDetectorsDescriptors}            & \textbf{77.1} & \textbf{82.8} & \textbf{91.9} & \underline{86.4} & \underline{90.7} & \underline{95.9} & 83.6 & 89.1 & 91.8 & 46.0 & 66.6 & \textbf{77.3} & \textbf{44.0} & \textbf{62.9} & \textbf{68.0}  \\
  ~~DenseVLAD~\cite{ToriiCVPR15PlaceRecognitionByViewSynthesis}+R2D2~\cite{RevaudNIPS19R2D2ReliableRepeatableDetectorsDescriptors}       & \underline{76.4} & 81.9 & 90.7 & \textbf{87.2} & \textbf{91.7} & \textbf{97.0} & 85.7 & 91.0 & 93.6 & \textbf{46.5} & \textbf{67.0} & \underline{76.8} & \underline{43.8} & 61.5 & 65.8  \\
  ~~NetVLAD~\cite{ArandjelovicCVPR16NetVLADPlaceRecognition}+R2D2~\cite{RevaudNIPS19R2D2ReliableRepeatableDetectorsDescriptors}         & 76.2 & 82.1 & 90.7 & 86.2 & 90.3 & 95.2 & \textbf{88.0} & \textbf{93.2} & \textbf{95.4} & 44.0 & 62.7 & 72.7 & \underline{43.8} & \underline{62.6} & \underline{67.6}  \\
  \hline
  \textit{ESAC~\cite{Brachmann2019ICCVa}} & &&&&&&&&&&&&&& \\

  ~~1 expert      & 0.0 & 0.0 & 0.5 & 0.0 & 0.4 & 9.1 & 0.0 & 0.5 & 4.8 & 0.0 & 0.0 & 0.0 & 0.0 & 0.0 & 0.0  \\
  ~~10 experts    & 0.9 & 2.9 & 6.5 & 21.1 & 46.0 & 62.1 & 20.1 & 39.5 & 62.2 & 4.8 & 17.9 & 35.8 & 5.9 & 19.2 & 33.9  \\
  ~~20 experts    & 1.7 & 3.8 & 7.8 & 34.9 & 62.6 & 78.0 & 30.5 & 47.7 & 65.7 & 7.5 & 22.0 & 38.1 & 9.3 & 23.6 & 37.5  \\
  ~~50 experts    & 3.5 & 8.2 & 12.6 & 43.3 & 66.3 & 77.0 & 45.2 & 62.5 & 73.1 & 11.4 & 31.1 & 44.7 & 11.2 & 26.1 & 39.9  \\
  \hline
  \textit{PoseNet~\cite{KendallICCV15PoseNetCameraRelocalization}}              & 0.0 & 0.0 & 0.1 & 0.0 & 0.0 & 0.4 & 0.0 & 0.0 & 0.1 & 0.0 & 0.0 & 0.1 & 0.0 & 0.0 & 0.0  \\
\hline
\end{tabular}
}
 
 \vspace{-0.3cm}
 
\caption{Results of various visual localization methods on the 5 NAVER LABS datasets, with the percentages of successfully localized test images within three thresholds for each datasets. The best method is shown in bold, the second best is underlined.}

\vspace{-0.5cm}

\label{tab:testres}
\end{table*}

In this section, we benchmark state-of-the-art visual localization algorithms on the proposed datasets.
We selected relevant approaches representing the different algorithmic groups namely structure, image retrieval, scene coordinate regression, and absolute pose regression based methods.

\subsection{Evaluated methods}
\label{sec:eval:methods}

\noindent \textbf{Structure-based methods} use an SFM model~\cite{Sattler2011ICCV,Snavely08IJCV,Heinly2015CVPR,SchonbergerCVPR16StructureFromMotionRevisited,sfm_survey_2017} as a representation of the environment in which new images can be registered in order to localize them. 
This is done with local feature matching using descriptors such as SIFT~\cite{Lowe04IJCV}, R2D2~\cite{RevaudNIPS19R2D2ReliableRepeatableDetectorsDescriptors, RevaudX19R2D2ReliableRepeatableDetectorsDescriptors}, D2-Net~\cite{DusmanuCVPR19D2NetDeepLocalFeatures} or SuperPoint~\cite{AndersonCVPR18SuperpointInterestPoint,Sarlin2020CVPR}. 
Image retrieval using global image representations such as DenseVLAD~\cite{ToriiCVPR15PlaceRecognitionByViewSynthesis}, NetVLAD~\cite{ArandjelovicCVPR16NetVLADPlaceRecognition}, APGeM~\cite{RevaudICCV19LearningwithAPTrainingImgRetrievalListwiseLoss}, or DELG~\cite{CaoX20UnifyingDeepLocalGlobalFeatures} can be used to select the image pairs to match~\cite{Torii2019TPAMI,Taira2019TPAMI,SarlinCVPR19FromCoarsetoFineHierarchicalLocalization} rather than brute force matching of the entire reconstruction.
Structure-based methods perform very well on a large variety of datasets but come with the burden of constructing and maintaining large 3D maps.

For our experiments, we chose a custom pipeline consisting of two stages: mapping and localization. 
For mapping, we start by selecting the image pairs to match using the global image representations mentioned above, we then extract the local keypoints and match them using cross validation and geometric verification, and finally, we compute the 3D points using COLMAP's~\cite{schoenberger2016sfm} point triangulator.
For localization, for each query image we first retrieve the 20 most similar database (map) images, we then extract the local keypoints and match the 20 image pairs again using cross validation and geometric verification. 
Since the database images are associated with 3D points from the map, we can use the resulting 2D-3D matches to compute the camera pose. 
For this, we use COLMAP's image triangulator.
We also evaluated a traditional pipeline, fully based on COLMAP~\cite{schoenberger2016sfm}, with SIFT~\cite{Lowe04IJCV} features and vocabulary tree based matching.
For both pipelines, we use less strict bundle adjustment parameters to maximize the percentage of localized images, even if this means that some of them might be significantly wrong (see the supplementary material for these parameters). 
We denote as \textit{GLOBAL}+\textit{LOCAL} the method using \textit{GLOBAL} features for retrieval and \textit{LOCAL} features for matching.

\noindent \textbf{Scene point regression methods} establish the 2D-3D correspondences between image pixel locations and the 3D space using random forests~\cite{ShottonCVPR13SceneCoordinateRegression,Massiceti2017ICRA} or deep neural networks~\cite{BrachmannCVPR18LearningLessIsMore6DLocalization}. 
The latter (notably DSAC++~\cite{BrachmannCVPR18LearningLessIsMore6DLocalization}) report high pose accuracy on small scale environments but fail when scaled up to larger scenes~\cite{Weinzaepfel_2019_CVPR}. 
To overcome this limitation, \cite{Brachmann2019ICCVa} proposes ESAC.
The idea is to use the mixture of experts strategy to separate large scenes into smaller scenes, called experts. 
A gating network is used to first select a subset of experts, and second to apply each of them (\ie, a DSAC++ like network) to estimate pose hypotheses. 
The pose hypothesis with maximum sample consensus is selected as final pose.
Similarly, \cite{Li_2020_CVPR} proposes a coarse-to-fine strategy within the neural network to increase the size of environments scene point regression can be successfully used.
For large-scale datasets however, \cite{Li_2020_CVPR} suggests to use image retrieval as additional conditioning.

For our experiments, we chose ESAC~\cite{Brachmann2019ICCVa} with various numbers of experts (1, 10, 20 and 50).
Note that using 1 expert is similar to DSAC++~\cite{BrachmannCVPR18LearningLessIsMore6DLocalization} (minor differences are listed in~\cite{Brachmann2019ICCVa}).
To initialize the experts, pseudo-depth maps are required and are generated from an SFM model, here we use APGeM+R2D2. This is simply used to initialize the experts, they are then refined using a reprojection loss which does not depend on these pseudo depth maps.

\noindent \textbf{Absolute pose regression methods} use deep neural networks to directly estimate the 6DoF pose from an input image~\cite{KendallICCV15PoseNetCameraRelocalization,LaskarICCVWS17CameraRelocalizationRelativePosesCNN,BalntasECCV18RelocNetMetricLearningRelocalisation,SattlerCVPR19UnderstandingLimitationsPoseRegression}. 
Similar to scene point regression, these methods do not scale up well to large scenes and perform similar to pose approximation methods~\cite{SattlerCVPR19UnderstandingLimitationsPoseRegression}.
For our experiments, we chose PoseNet~\cite{KendallICCV15PoseNetCameraRelocalization}.
We follow~\cite{HenriquesCVPR18MapNetAllocentricSpatialMemory} and train the model for 300 epochs with Adam.

\subsection{Evaluation metric}

Following the well established evaluation protocol from~\cite{SattlerCVPR18Benchmarking6DoFOutdoorLoc}, we compute the percentage of successfully localized images within three position and orientation thresholds representing high (0.1m, 1$\degree$), medium (0.25m, 2$\degree$), and low (1m, 5$\degree$) accuracy. We chose the highest position accuracy being 10cm because, according to our experience, this is the minimum accuracy necessary for real-world applications. 

\subsection{Discussion}
\label{sec:eval:discussion}

Table~\ref{tab:testres} shows the results of state-of-the-art visual localization methods on the Galaxy images from the test set of the 5 NAVER LABS datasets. We also provide the results on other splits (validation set and/or Basler images) in the supplementary material, as well as plots showing the percentage of localized images for varying thresholds.

First, we observe that structure-based methods significantly outperform coordinate point regression and absolute pose regression approaches.
For instance, DELG+R2D2 successfully localizes more than 75\% (resp.~90\%) of the images at the high (resp. low) accuracy thresholds on the 3 Dept.~datasets, while ESAC performs worse (below 50\% at the high accuracy thresholds), and PoseNet simply fails.
The performance on the 2 Metro St.~datasets is lower but the overall ranking remains similar. 
Hypotheses for the lower performance on Metro St.~include the narrower scene, the crowdedness, the symmetry of the scene (in B2, see Figure~\ref{fig:Gangnam}), and the screens changing their content regularly.

In terms of global features for image retrieval, DELG, DenseVLAD, and NetVLAD have overall similar performance, APGeM performing slightly worse, confirming the finding of~\cite{Pion3DV20BenchmarkingImageRetrievalVisualLocalization} for the indoor dataset~\cite{Sun2017CVPR}. For local features, R2D2 often performs slightly better than D2-Net and SIFT, confirming its robustness to visual localization challenges reported in \cite{Pion3DV20BenchmarkingImageRetrievalVisualLocalization,RevaudNIPS19R2D2ReliableRepeatableDetectorsDescriptors}.

For scene point regression methods, the performance with 1 expert is low. This is not surprising as it is equivalent to DSAC++ which is known to not scale to large environments~\cite{Weinzaepfel_2019_CVPR,Li_2020_CVPR}. Using more experts allow to better handle larger datasets like ours, but the performance remains behind structure-based methods. Further increasing the number of experts does not necessarily increase the performance due to the unbalance of the clustering applied to 3D points to separate the experts' areas. The performance of ESAC is particularly low on Dept.~B1, which can be led back to the blur in the images due to the low lighting conditions and the repetitive textures as well as the dynamics in food court.

\begin{table}
 \centering
 \resizebox{\linewidth}{!}{
 \begin{tabular}{lrrr}
 \hline
  Test set       & \# low freq. & performance for & performance for \\
 Galaxy images  & images      & low freq. images  & other images \\
 \hline
Dept. B1 &        96/2539   &   41.7  &   84.4 \\
Dept. 1F &        43/975    &   74.4  &   91.4 \\
Dept. 4F &        520/2411  &   67.1  &   95.2 \\
Metro St. B1   &        217/4334  &   49.8  &   67.5 \\
Metro St. B2   &        153/1786  &   49.7  &   64.2 \\
\hline
 \end{tabular}
}
 
 \vspace{-0.2cm}
 
 \caption{Impact of the absence of high frequencies (low freq.~images) for DELG+R2D2. We report the percentage of successfully localized images within 25cm and 2\degree.}
 \label{tab:resblur}
 
 \vspace{-0.3cm}
 
\end{table}

Absolute pose regression does not perform well at all, due to limitations of such approaches, see~\cite{SattlerCVPR19UnderstandingLimitationsPoseRegression}.

In the supplementary material we report results for Basler images and we obtain significantly higher performances. This is not surprising since these images are taken with the same cameras as the training images. In particular, we observe that the performance of ESAC on the Basler test set significantly increases on 4 out of 5 datasets. This suggests that scene point regression might not be sufficiently robust to varying cameras between training and test time. 
We also report the performance on the validation set in the supplementary material and draw the same conclusion.

To better evaluate the impact of some specific challenges, in Table~\ref{tab:resblur} and Table~\ref{tab:rescrowd}, we report DELG+R2D2 performance when considering only images without high frequency content (\ie, blurry and/or with large textureless area) and crowded images, respectively.
We define images without high frequency, following~\cite{renting2008image,bluriness}: we compute the mean absolute difference between the original image and its version without high frequencies, which is done in the Fourier domain, and threshold it at 20.
From Table~\ref{tab:resblur}, we observe that the performance is significantly worse for low frequency images, for instance from 95.2\% successfully localized images within 25cm and 2\degree drops down to 67.1\% on Dept.~4F, the dataset with the most low frequency images due to large textureless temporary walls on a construction site. 
The main reason for this decrease in performance is the fact that local features used for matching mainly rely on high frequency signals both for detecting and describing the keypoints.
For analyzing crowdedness, we use images where at least 20\% of the pixels belong to the human class, according to a semantic segmentation approach. The results reported in Table~\ref{tab:rescrowd} show that it also represents a significant challenge. 
For instance, on Metro St.~B1 (largest number of crowded images) we observe a drop of performance (successfully localized images within 25cm and 2\degree) from 69.1\% to 50.1\%. This is due to the fact that humans introduce irrelevant data both for image retrieval and local feature matching.
Despite these two challenges, the performance of structure-based method remain quite high, with about 50\% of the images successfully localized within 25cm and 2\degree.

\begin{table}
 \centering
 \resizebox{\linewidth}{!}{
 \begin{tabular}{lrrr}
 \hline
 Test set        & \# crowded   & performance for & performance for \\
 Galaxy images  &  images      & crowded images  & other images \\
 \hline
Dept. B1          & 35/2539   &   77.1  &   82.9 \\
Dept. 1F          & 6/ 975    & 100.0   &  90.6 \\
Dept. 4F          & 10/2411   &   70.0   &  89.3 \\
Metro St. B1            & 565/4334    &  50.1    & 69.1 \\
Metro St. B2            & 169/1786    &  56.2    & 63.6 \\
\hline
 \end{tabular}
}
 \vspace{-0.2cm}
 \caption{Impact of crowdedness for DELG+R2D2. 
We report the performance as the percentage of successfully localized images within 25cm and 2\degree.}
 \vspace{-0.3cm}
 \label{tab:rescrowd}
\end{table}

\section{Conclusion}
We introduce the NAVER LABS datasets that are first to propose large-scale densely-sampled indoor environments with challenges like dynamic objects, moving crowds, and changing scenes.
We capture the data with a specific mapping platform equipped with LiDARs, cameras, and wheel odometry and propose a new pipeline based on LiDAR SLAM and SFM to generate ground truth poses. Finally, we provide a benchmark of existing methods on these datasets.
We believe the NAVER LABS datasets will foster new research directions in visual localization, image retrieval, and local feature extraction and may also be used for other tasks such as depth estimation or completion, LiDAR place recognition, and map change detection.

\newpage
\appendix

\section{Appendix}

In Section~\ref{sec:details} we first provide more details about the structure-based methods we benchmark on our datasets.
Then, in Section~\ref{sec:results}, we provide results on the test sets using Basler cameras as well as results on the validation sets using both, Basler and Galaxy cameras.
Furthermore, for the Galaxy test sets we present plots with varying accuracy thresholds.
Finally, to provide more insights on our datasets, we show the 4 best and 4 worst localized images.

\begin{figure*}
 \includegraphics[width=0.32\linewidth]{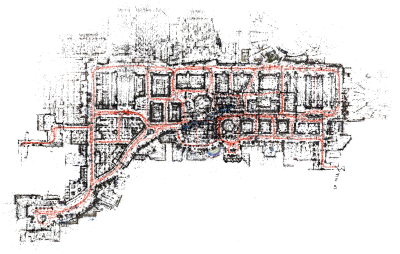}
 \hspace{0.01\linewidth}
 \includegraphics[width=0.32\linewidth]{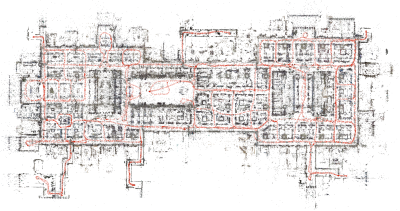}
 \hspace{0.01\linewidth}
 \includegraphics[width=0.32\linewidth]{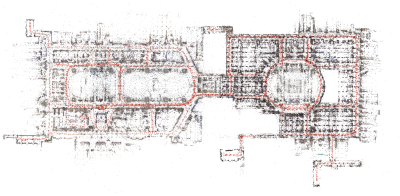}
 \\
 \includegraphics[width=0.32\linewidth]{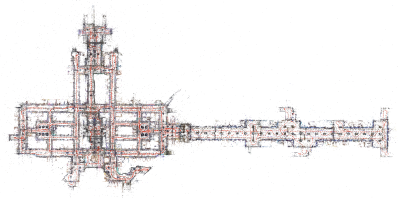}
 \hspace{0.01\linewidth}
 \includegraphics[width=0.32\linewidth]{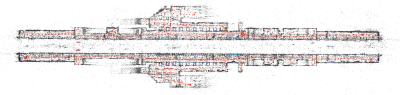}
 \hspace*{0.16\linewidth}
 \caption{SFM reconstructions of our datasets (camera centers in red). From left to right: Top: Dept.~B1, Dept.~1F, Dept.~4F, Bottom: Metro St.~B1, Metro St.~B2}
 \label{fig:sfm_models}
\end{figure*}

\begin{figure*}
 \includegraphics[width=0.32\linewidth]{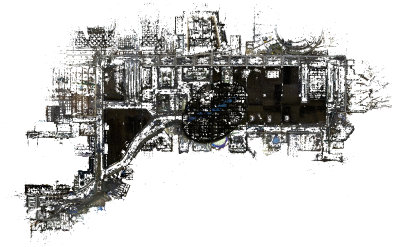}
 \hspace{0.01\linewidth}
 \includegraphics[width=0.32\linewidth]{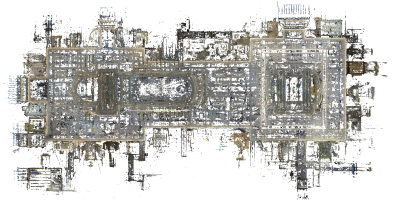}
 \hspace{0.01\linewidth}
 \includegraphics[width=0.32\linewidth]{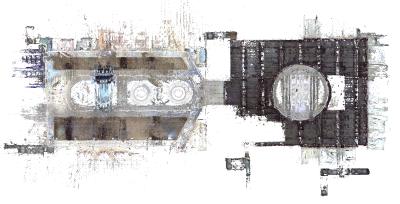}
 \\
 \includegraphics[width=0.32\linewidth]{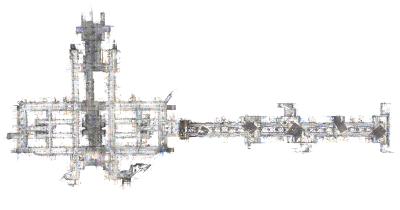}
 \hspace{0.01\linewidth}
 \includegraphics[width=0.32\linewidth]{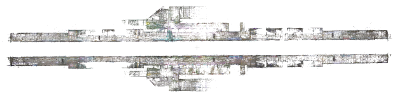}
 \hspace*{0.16\linewidth}
 \caption{Dense reconstructions of our datasets. From left to right: Top: Dept.~B1, Dept.~1F, Dept.~4F, Bottom: Metro St.~B1, Metro St.~B2}
 \label{fig:mvs_models}
\end{figure*}

\section{Details of the structure-based methods}
\label{sec:details}

For our experiments, we chose a custom pipeline consisting of two stages: mapping and localization.
For both, we used different selections of global and local representations.
After extracting the local keypoints and robustly matching them using cross validation and geometric verification, we use the \textsf{point\_triangulator}
of COLMAP\footnote{\url{https://colmap.github.io}} to compute the 3D point locations for mapping
and the \textsf{image\_registrator} of COLMAP to compute the query camera pose from a set of 2D-3D matches.
In Table~\ref{tab:params}, we show the COLMAP parameters we used (taken from~\cite{localfeateval}).
We also evaluated a traditional pipeline, fully based on COLMAP~\cite{schoenberger2016sfm}, with SIFT~\cite{Lowe04IJCV} features and vocabulary tree based matching.
Since we already obtained the camera intrinsic parameters during the calibration process described in Section~\ref{sec:processing}, we fix them for the localization experiments.
Figure~\ref{fig:sfm_models}~and~\ref{fig:mvs_models} show SFM models and dense reconstructions of our datasets.

\section{Further benchmark results}
\label{sec:results}

We denote the experiments as \textit{GLOBAL}+\textit{LOCAL}, where we use \textit{GLOBAL} features to generate image pairs used during mapping and localization, and \textit{LOCAL} features for keypoint matching (\eg DenseVLAD+D2-Net).
See Section~\ref{sec:eval:methods} for more details about the method.

In Table~\ref{tab:testres}, we have shown the results obtained with various modern visual localization algorithms on the Galaxy images from the test set.
In Table~\ref{tab:big}, we provide the results for the same methods on the Basler images from the test set, as well as Galaxy and Basler images from the validation set.
As before, we report the percentages of successfully localized images for three thresholds, high (0.1m, 1$\degree$), medium (0.25m, 2$\degree$), and low (1m, 5$\degree$) accuracy.

\begin{figure*}
 \includegraphics[width=0.32\linewidth]{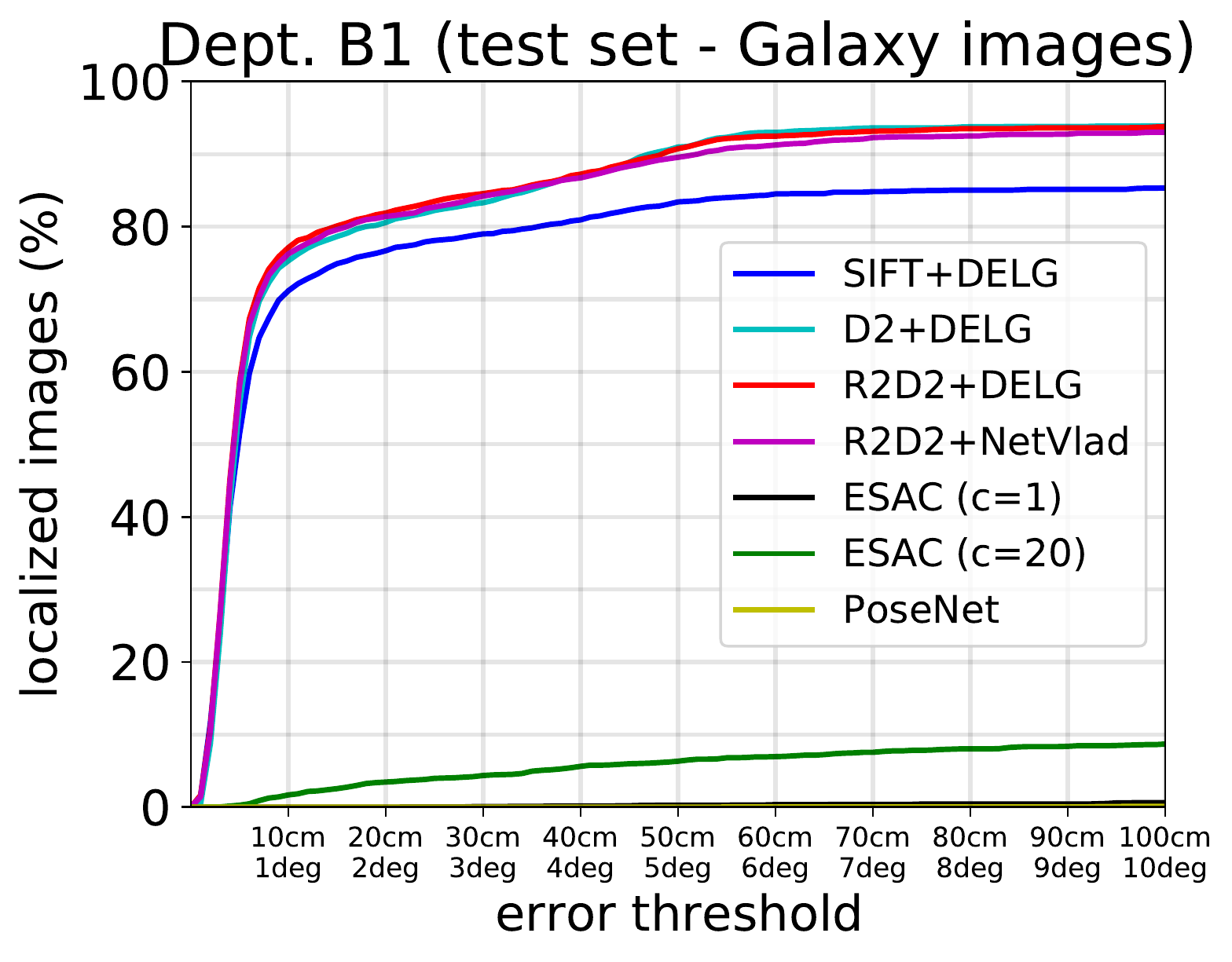}
 \hspace{0.01\linewidth}
 \includegraphics[width=0.32\linewidth]{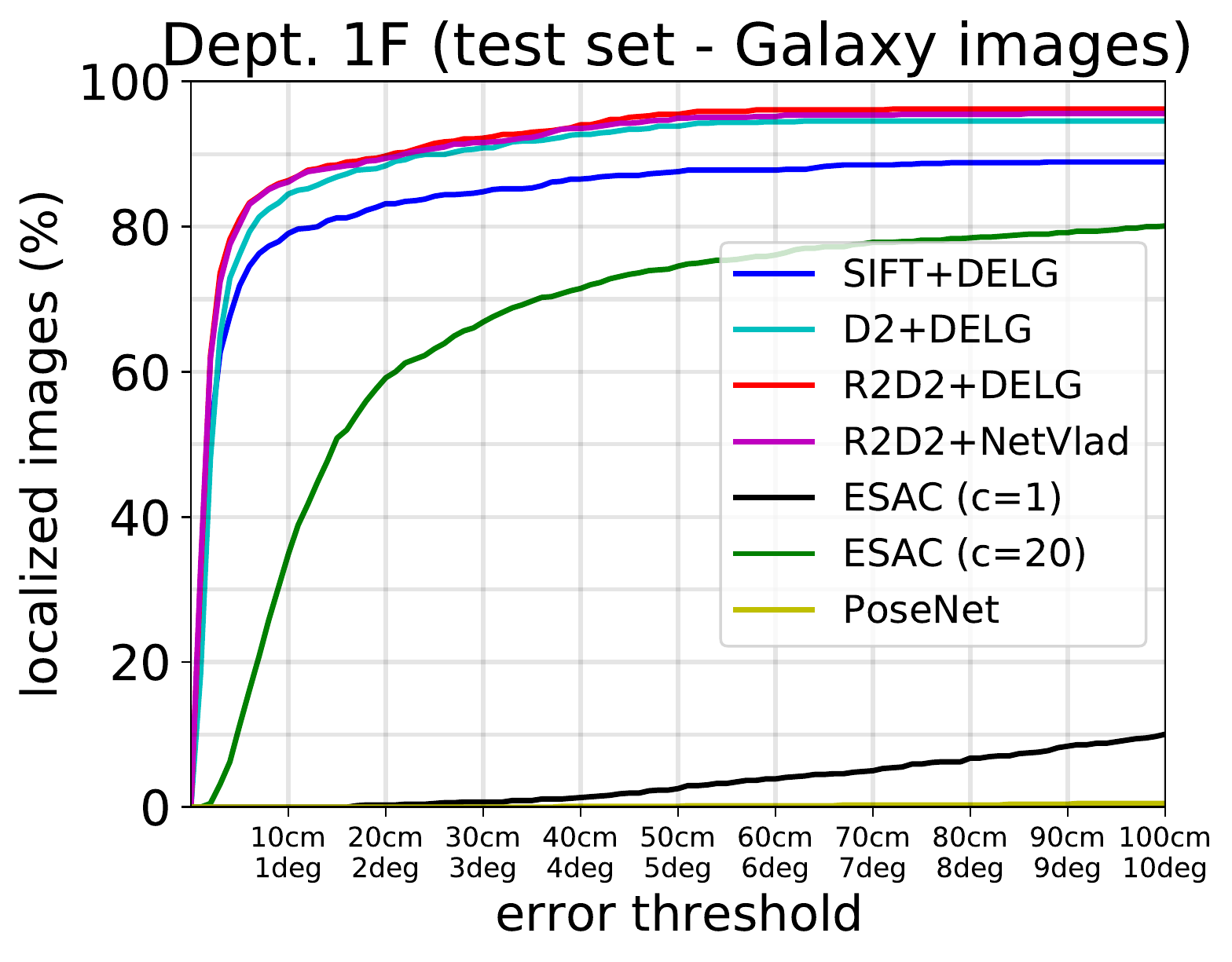}
 \hspace{0.01\linewidth}
 \includegraphics[width=0.32\linewidth]{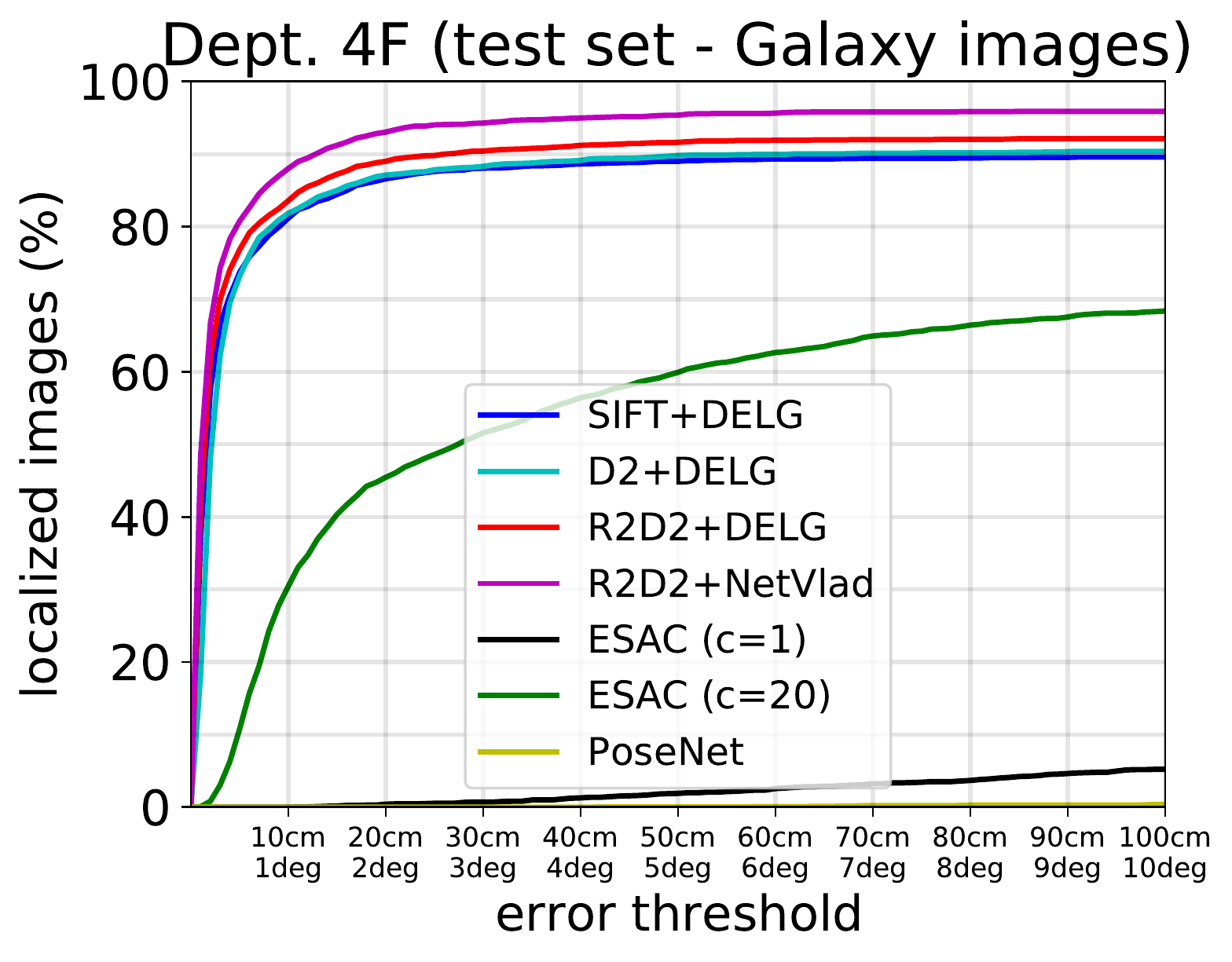}
 \\
 \hspace*{0.16\linewidth}
  \includegraphics[width=0.32\linewidth]{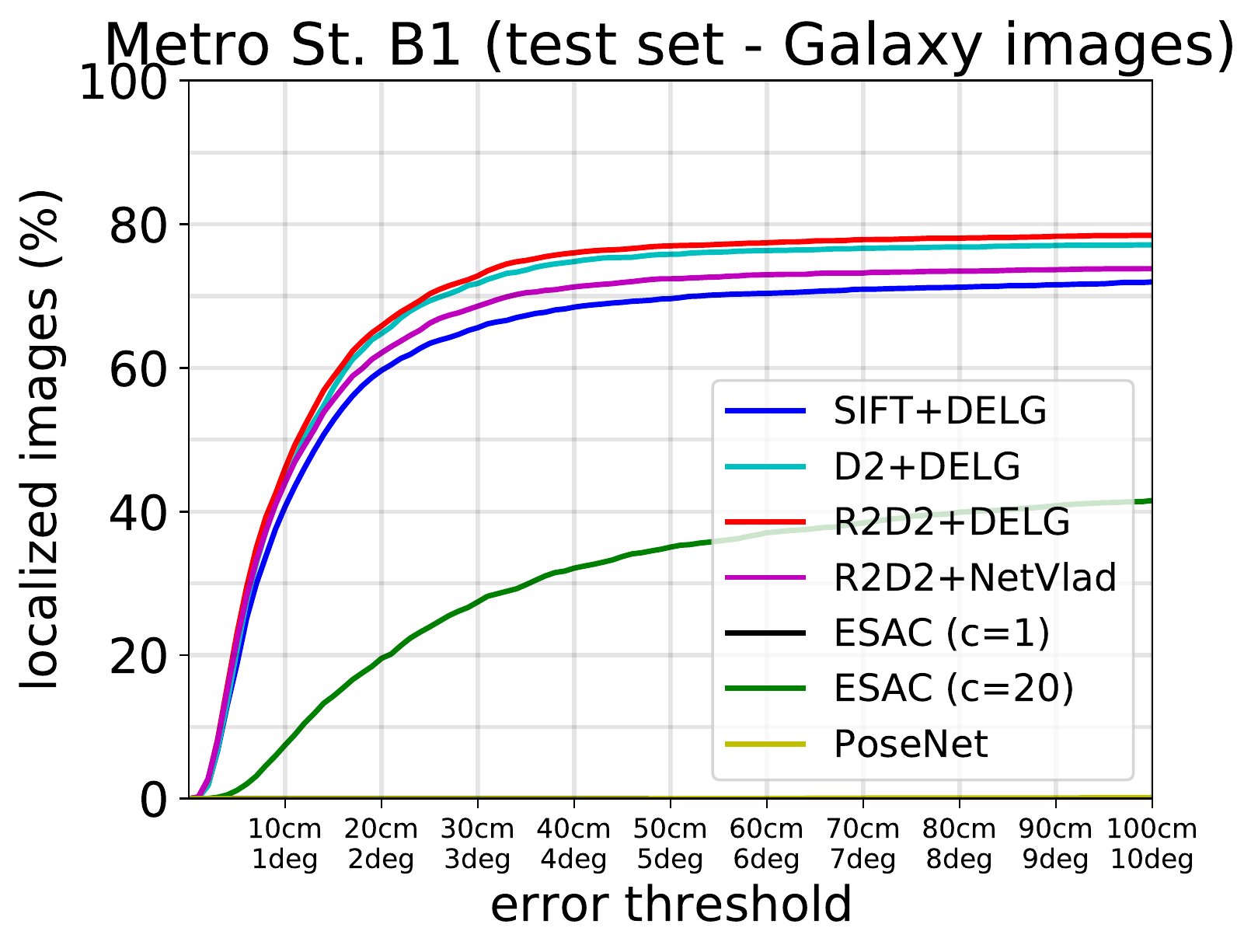}
 \hspace{0.01\linewidth}
 \includegraphics[width=0.32\linewidth]{images/benchmark_plots/GangnamStation_B1_test_galaxy.pdf}
 \hspace*{0.16\linewidth}
 \caption{Results with varying error threshold on Galaxy images for all 5 NAVER LABS localization datasets. The angular error threshold in degree varies as 10 times the positional error threshold in cm.}
 \label{fig:plots}
\end{figure*}

In addition, for the Galaxy test sets, in Figure~\ref{fig:plots} we show the percentages of successfully localized images for some of the methods when varying the thresholds for the positional error between 0 and 1m
with an angular error threshold in degree varying as 10 times the positional error threshold in cm, \ie, (10cm, 1\degree), (20cm, 2\degree), etc.
These plots visually illustrate the comparison between different localization methods and confirm the observations drawn from the tables:
\begin{itemize}
\item The results for the Basler images are overall better than the ones for the Galaxy images both, on the test as well as on the validation sets.
This is not surprising as the difference in quality and resolution between the images from the industrial Basler cameras and the ones from the smartphone cameras introduces a domain bias between image representations (local and global).
Since for mapping we only use Basler cameras, this domain bias affects the localization results of all methods for all datasets.
\item Concerning the comparison between different methods, in these tables we observe similar behaviour as in the table shown in the main paper. Hence they confirm the observations discussed in Section~\ref{sec:eval:discussion} of the main paper, namely that the structure-based methods significantly outperform ESAC, and that PoseNet completely fails to localize the query images. Furthermore, we have similar performance when varying the global image representation for retrieval, without having a clear winning representation. Finally, for local features, R2D2 slightly outperforms D2-Net and both yield much better localization results than using SIFT.
\item Finally, there is little difference between the results on validation and test sets, showing that the difficulty between the two zones remains similar and hence makes the validation set appropriate to be used for algorithm development, parameter optimization, and model tuning.
\end{itemize}

\begin{table}[]
\center
\resizebox{0.9\linewidth}{!}{
\begin{tabular}{|l|c|c|}
\hline
\textbf{COLMAP parameters} & \textbf{triangulator} & \textbf{registrator} \\ \hline \hline
Mapper.ba\_refine\_focal\_length & 0 & 0 \\ \hline
Mapper.ba\_refine\_principal\_point & 0 & 0 \\ \hline
Mapper.ba\_refine\_extra\_params & 0 & 0 \\ \hline
Mapper.min\_num\_matches & 15 & 4 \\ \hline
Mapper.init\_min\_num\_inliers & 100 & 4 \\ \hline
Mapper.abs\_pose\_min\_num\_inliers & 30 & 4 \\ \hline
Mapper.abs\_pose\_min\_inlier\_ratio & 0.25 & 0.05 \\ \hline
Mapper.ba\_local\_max\_num\_iterations & 25 & 50 \\ \hline
Mapper.abs\_pose\_max\_error & 12 & 20 \\ \hline
Mapper.filter\_max\_reproj\_error & 4 & 12 \\ \hline
\end{tabular}
}
\vspace{0.1cm}
\caption{The parameters we used for COLMAP~\cite{schoenberger2016sfm}.}
\label{tab:params}
\end{table}

\begin{table*}
\resizebox{\linewidth}{!}{
\begin{tabular}{|l|rrr|rrr|rrr|rrr|rrr|}
\hline
\textbf{Test set - \emph{Basler} images} & \multicolumn{3}{c|}{ Dept. B1 } & \multicolumn{3}{c|}{ Dept. 1F } & \multicolumn{3}{c|}{ Dept. 4F } & \multicolumn{3}{c|}{ Metro St. B1 } & \multicolumn{3}{c|}{ Metro St. B2 }  \\
\textbf{Algorithm / Accuracy th.} & 0.1m,1\degree   & 0.25m,2\degree  & 1m,5\degree & 0.1m,1\degree   & 0.25m,2\degree  & 1m,5\degree & 0.1m,1\degree   & 0.25m,2\degree  & 1m,5\degree & 0.1m,1\degree   & 0.25m,2\degree  & 1m,5\degree & 0.1m,1\degree   & 0.25m,2\degree  & 1m,5\degree \\
\hline
\textit{structure-based methods} &&&&&&&&&&&&&&& \\
  ~~APGeM~\cite{RevaudICCV19LearningwithAPTrainingImgRetrievalListwiseLoss}+SIFT~\cite{Lowe04IJCV}         & 87.9 & 91.5 & 92.8 & 85.8 & 87.4 & 89.0 & 92.8 & 94.0 & 94.5 & 67.6 & 75.8 & 78.8 & 69.2 & 72.3 & 73.8  \\
  ~~DELG~\cite{CaoX20UnifyingDeepLocalGlobalFeatures}+SIFT~\cite{Lowe04IJCV}          & 88.6 & 92.0 & 93.3 & 83.7 & 85.4 & 86.6 & 92.7 & 93.8 & 94.1 & 67.0 & 74.7 & 77.8 & 69.0 & 71.9 & 73.8  \\
  ~~DenseVLAD~\cite{ToriiCVPR15PlaceRecognitionByViewSynthesis}+SIFT~\cite{Lowe04IJCV}     & 91.0 & 94.0 & 95.0 & 87.0 & 88.3 & 89.6 & 92.2 & 93.4 & 94.0 & 68.5 & 75.7 & 78.3 & 69.0 & 71.9 & 73.8  \\
  ~~NetVLAD~\cite{ArandjelovicCVPR16NetVLADPlaceRecognition}+SIFT~\cite{Lowe04IJCV}       & 90.5 & 93.6 & 94.8 & 85.9 & 87.4 & 89.3 & 93.2 & 94.2 & 94.6 & 65.6 & 72.9 & 75.6 & 70.5 & 73.5 & 75.0  \\
  ~~SIFT~\cite{Lowe04IJCV}+vocab.~tree (COLMAP~\cite{schoenberger2016sfm}) & 77.1 & 84.0 & 85.9 & 80.1 & 82.1 & 83.4 & 86.0 & 87.1 & 87.6 & 63.6 & 71.8 & 75.8 & 65.8 & 69.1 & 70.5  \\
  ~~APGeM~\cite{RevaudICCV19LearningwithAPTrainingImgRetrievalListwiseLoss}+D2-Net~\cite{DusmanuCVPR19D2NetDeepLocalFeatures}           & 92.4 & 95.6 & 96.2 & 91.6 & 92.3 & 92.7 & 92.9 & 94.0 & 94.4 & 70.8 & 78.8 & 81.3 & 67.2 & 71.2 & 73.1  \\
  ~~DELG~\cite{CaoX20UnifyingDeepLocalGlobalFeatures}+D2-Net~\cite{DusmanuCVPR19D2NetDeepLocalFeatures}            & 93.2 & 95.7 & 96.3 & 90.5 & 91.9 & 92.3 & 92.5 & 93.7 & 94.3 & 70.1 & 78.4 & 81.1 & 66.6 & 70.5 & 72.2  \\
  ~~DenseVLAD~\cite{ToriiCVPR15PlaceRecognitionByViewSynthesis}+D2-Net~\cite{DusmanuCVPR19D2NetDeepLocalFeatures}       & 94.7 & 96.6 & 97.0 & 90.8 & 91.9 & 92.5 & 92.2 & 93.6 & 94.2 & 70.5 & 77.9 & 79.8 & 66.8 & 70.1 & 71.5  \\
  ~~NetVLAD~\cite{ArandjelovicCVPR16NetVLADPlaceRecognition}+D2-Net~\cite{DusmanuCVPR19D2NetDeepLocalFeatures}         & 94.6 & 96.7 & 97.1 & 91.7 & 93.1 & 93.5 & 92.8 & 93.9 & 94.5 & 68.5 & 75.7 & 78.0 & 68.7 & 72.4 & 73.8  \\
  ~~APGeM~\cite{RevaudICCV19LearningwithAPTrainingImgRetrievalListwiseLoss}+R2D2~\cite{RevaudNIPS19R2D2ReliableRepeatableDetectorsDescriptors}         & 93.8 & 96.2 & 96.7 & \underline{93.1} & \underline{94.1} & \underline{94.8} & \underline{93.8} & \underline{94.8} & \textbf{95.2} & \underline{73.2} & \textbf{80.9} & \textbf{83.3} & \underline{71.6} & \underline{73.9} & \underline{75.5}  \\
  ~~DELG~\cite{CaoX20UnifyingDeepLocalGlobalFeatures}+R2D2~\cite{RevaudNIPS19R2D2ReliableRepeatableDetectorsDescriptors}          & 94.3 & 96.6 & 97.0 & 92.3 & 93.3 & 93.9 & 93.5 & 94.7 & \underline{95.1} & \textbf{73.3} & \underline{80.6} & \underline{82.8} & 71.0 & 73.3 & 74.7  \\
  ~~DenseVLAD~\cite{ToriiCVPR15PlaceRecognitionByViewSynthesis}+R2D2~\cite{RevaudNIPS19R2D2ReliableRepeatableDetectorsDescriptors}     & \textbf{95.6} & \underline{97.2} & \underline{97.6} & 91.8 & 93.0 & 93.4 & 93.5 & 94.5 & 94.9 & 73.1 & 80.2 & 82.1 & 71.1 & 73.3 & 74.6  \\
  ~~NetVLAD~\cite{ArandjelovicCVPR16NetVLADPlaceRecognition}+R2D2~\cite{RevaudNIPS19R2D2ReliableRepeatableDetectorsDescriptors}       & \underline{95.4} & \textbf{97.4} & \textbf{97.7} & \textbf{93.2} & \textbf{94.6} & \textbf{95.1} & \textbf{93.9} & \textbf{94.8} & 95.1 & 71.1 & 78.1 & 80.3 & \textbf{72.9} & \textbf{75.3} & \textbf{76.8}  \\
\hline
\textit{ESAC~\cite{Brachmann2019ICCVa}} &&&&&&&&&&&&&&& \\
  ~~1 expert           & 0.0 & 0.0 & 0.6 & 0.0 & 0.0 & 0.9 & 0.7 & 6.0 & 22.7 & 0.0 & 0.0 & 0.0 & 0.0 & 0.0 & 0.0  \\
  ~~10 experts         & 0.6 & 4.1 & 13.4 & 2.9 & 8.8 & 17.5 & 64.8 & 82.5 & 88.7 & 30.6 & 55.1 & 68.9 & 30.8 & 50.9 & 65.3  \\
  ~~20 experts         & 4.5 & 13.5 & 25.8 & 10.3 & 21.9 & 34.1 & 71.9 & 84.9 & 89.3 & 36.4 & 60.7 & 72.6 & 35.7 & 52.2 & 62.6  \\
  ~~50 experts         & 12.9 & 25.2 & 37.0 & 16.6 & 30.6 & 40.3 & 76.8 & 86.1 & 90.0 & 47.3 & 67.5 & 75.5 & 35.6 & 53.7 & 63.0  \\
\hline
  \textit{PoseNet~\cite{KendallICCV15PoseNetCameraRelocalization}}     & 0.0 & 0.0 & 0.3 & 0.0 & 0.0 & 0.0 & 0.0 & 0.0 & 0.2 & 0.0 & 0.0 & 0.2 & 0.0 & 0.0 & 0.0  \\
\hline
\multicolumn{16}{c}{~~} \\
\hline
\textbf{Validation set - \emph{Galaxy} images} & \multicolumn{3}{c|}{ Dept. B1 } & \multicolumn{3}{c|}{ Dept. 1F } & \multicolumn{3}{c|}{ Dept. 4F } & \multicolumn{3}{c|}{ Metro St. B1 } & \multicolumn{3}{c|}{ Metro St. B2 }  \\
\textbf{Algorithm / Accuracy th.} & 0.1m,1\degree   & 0.25m,2\degree  & 1m,5\degree & 0.1m,1\degree   & 0.25m,2\degree  & 1m,5\degree & 0.1m,1\degree   & 0.25m,2\degree  & 1m,5\degree & 0.1m,1\degree   & 0.25m,2\degree  & 1m,5\degree & 0.1m,1\degree   & 0.25m,2\degree  & 1m,5\degree \\
\hline
\textit{structure-based methods} &&&&&&&&&&&&&&& \\
  ~~APGeM~\cite{RevaudICCV19LearningwithAPTrainingImgRetrievalListwiseLoss}+SIFT~\cite{Lowe04IJCV}         & 64.7 & 72.2 & 78.6 & 82.3 & 87.5 & 92.1 & 72.6 & 83.6 & 97.3 & 34.6 & 53.4 & 63.9 & 37.8 & 59.6 & 65.3  \\
  ~~DELG~\cite{CaoX20UnifyingDeepLocalGlobalFeatures}+SIFT~\cite{Lowe04IJCV}          & 64.0 & 70.1 & 77.3 & 83.5 & 89.2 & 94.5 & 71.9 & 84.8 & 98.1 & 35.2 & 53.0 & 64.2 & 39.2 & 60.9 & 66.2  \\
  ~~DenseVLAD~\cite{ToriiCVPR15PlaceRecognitionByViewSynthesis}+SIFT~\cite{Lowe04IJCV}     & 66.5 & 73.7 & 80.8 & 84.9 & 89.5 & 94.8 & 72.8 & 85.1 & 98.8 & 36.5 & 53.9 & 63.7 & 38.5 & 60.0 & 66.8  \\
  ~~NetVLAD~\cite{ArandjelovicCVPR16NetVLADPlaceRecognition}+SIFT~\cite{Lowe04IJCV}       & 66.9 & 73.4 & 80.8 & 82.6 & 89.1 & 94.1 & 71.9 & 84.6 & 98.1 & 31.5 & 47.5 & 56.6 & \underline{40.2} & 62.7 & \underline{68.2}  \\
  ~~SIFT~\cite{Lowe04IJCV}+vocab.~tree (COLMAP~\cite{schoenberger2016sfm}) & 64.2 & 71.6 & 77.3 & 82.7 & 87.1 & 93.5 & 72.6 & 84.9 & 98.5 & 33.0 & 49.3 & 59.6 & 31.4 & 50.7 & 55.8  \\
  ~~APGeM~\cite{RevaudICCV19LearningwithAPTrainingImgRetrievalListwiseLoss}+D2-Net~\cite{DusmanuCVPR19D2NetDeepLocalFeatures}           & 70.2 & 78.0 & 86.1 & 83.2 & 89.2 & 94.5 & 72.1 & 85.3 & 98.5 & 40.9 & 61.6 & 71.2 & 37.3 & 60.1 & 66.6  \\
  ~~DELG~\cite{CaoX20UnifyingDeepLocalGlobalFeatures}+D2-Net~\cite{DusmanuCVPR19D2NetDeepLocalFeatures}            & 69.7 & 76.5 & 87.2 & 85.7 & 90.3 & 95.9 & 72.6 & 85.8 & 98.6 & 41.6 & 61.8 & \underline{73.7} & 38.0 & 60.6 & 66.9  \\
  ~~DenseVLAD~\cite{ToriiCVPR15PlaceRecognitionByViewSynthesis}+D2-Net~\cite{DusmanuCVPR19D2NetDeepLocalFeatures}       & 70.7 & 77.2 & 87.1 & 85.0 & 89.8 & 95.1 & \underline{73.6} & \textbf{86.3} & 98.6 & 42.6 & 61.8 & 71.6 & 37.1 & 57.4 & 63.1  \\
  ~~NetVLAD~\cite{ArandjelovicCVPR16NetVLADPlaceRecognition}+D2-Net~\cite{DusmanuCVPR19D2NetDeepLocalFeatures}         & \underline{72.5} & \textbf{79.2} & \underline{88.5} & \underline{86.0} & 90.2 & 95.5 & \textbf{73.8} & \underline{86.0} & \underline{99.0} & 36.1 & 54.1 & 65.2 & 38.3 & 60.5 & 67.1  \\
  ~~APGeM~\cite{RevaudICCV19LearningwithAPTrainingImgRetrievalListwiseLoss}+R2D2~\cite{RevaudNIPS19R2D2ReliableRepeatableDetectorsDescriptors}         & 71.6 & 78.0 & 86.0 & 85.8 & 89.9 & 94.4 & 72.6 & 84.6 & 98.3 & 43.1 & 62.2 & 72.6 & 39.4 & 62.7 & 67.5  \\
  ~~DELG~\cite{CaoX20UnifyingDeepLocalGlobalFeatures}+R2D2~\cite{RevaudNIPS19R2D2ReliableRepeatableDetectorsDescriptors}          & 70.6 & 77.8 & 87.4 & \textbf{86.4} & \textbf{90.9} & \textbf{96.9} & 72.3 & 85.3 & 98.8 & \underline{43.4} & \underline{62.9} & \textbf{73.7} & 39.9 & \underline{63.1} & 68.0  \\
  ~~DenseVLAD~\cite{ToriiCVPR15PlaceRecognitionByViewSynthesis}+R2D2~\cite{RevaudNIPS19R2D2ReliableRepeatableDetectorsDescriptors}     & 71.9 & 77.8 & 87.9 & 85.8 & 90.5 & \underline{96.5} & 73.0 & 85.8 & \textbf{99.3} & \textbf{43.9} & \textbf{62.9} & 72.8 & 40.1 & 59.6 & 64.6  \\
  ~~NetVLAD~\cite{ArandjelovicCVPR16NetVLADPlaceRecognition}+R2D2~\cite{RevaudNIPS19R2D2ReliableRepeatableDetectorsDescriptors}       & \textbf{72.9} & \underline{79.0} & \textbf{89.2} & 85.7 & \underline{90.6} & 95.9 & 73.3 & 84.8 & \underline{99.0} & 39.0 & 56.4 & 66.8 & \textbf{41.3} & \textbf{63.5} & \textbf{68.7}  \\
\hline
\textit{ESAC~\cite{Brachmann2019ICCVa}} &&&&&&&&&&&&&&& \\
  ~~1 expert           & 0.0 & 0.0 & 0.1 & 0.0 & 0.1 & 3.0 & 0.0 & 0.5 & 9.9 & 0.0 & 0.0 & 0.0 & 0.0 & 0.0 & 0.0  \\
  ~~10 experts         & 0.6 & 1.8 & 6.0 & 17.3 & 46.7 & 73.2 & 28.1 & 60.5 & 86.8 & 1.1 & 6.6 & 17.2 & 4.4 & 14.8 & 22.9  \\
  ~~20 experts         & 2.3 & 5.7 & 10.8 & 33.8 & 61.9 & 81.2 & 45.4 & 70.2 & 89.2 & 4.1 & 13.8 & 26.8 & 4.3 & 13.4 & 22.5  \\
  ~~50 experts         & 5.4 & 9.1 & 14.2 & 49.7 & 71.5 & 84.1 & 45.2 & 69.9 & 85.1 & 7.9 & 20.3 & 32.7 & 6.0 & 16.1 & 24.6  \\
\hline
  \textit{PoseNet~\cite{KendallICCV15PoseNetCameraRelocalization}}     & 0.0 & 0.0 & 0.0 & 0.0 & 0.0 & 0.4 & 0.0 & 0.0 & 0.2 & 0.0 & 0.0 & 0.0 & 0.0 & 0.0 & 0.0  \\
\hline
\multicolumn{16}{c}{~~} \\
\hline
\textbf{Validation set - \emph{Basler} images} & \multicolumn{3}{c|}{ Dept. B1 } & \multicolumn{3}{c|}{ Dept. 1F } & \multicolumn{3}{c|}{ Dept. 4F } & \multicolumn{3}{c|}{ Metro St. B1 } & \multicolumn{3}{c|}{ Metro St. B2 }  \\
\textbf{Algorithm / Accuracy th.} & 0.1m,1\degree   & 0.25m,2\degree  & 1m,5\degree & 0.1m,1\degree   & 0.25m,2\degree  & 1m,5\degree & 0.1m,1\degree   & 0.25m,2\degree  & 1m,5\degree & 0.1m,1\degree   & 0.25m,2\degree  & 1m,5\degree & 0.1m,1\degree   & 0.25m,2\degree  & 1m,5\degree \\
\hline
\textit{structure-based methods} &&&&&&&&&&&&&&& \\
  ~~APGeM~\cite{RevaudICCV19LearningwithAPTrainingImgRetrievalListwiseLoss}+SIFT~\cite{Lowe04IJCV}         & 80.9 & 86.3 & 88.3 & 91.9 & 93.2 & 94.4 & 97.8 & 99.9 & \textbf{100.0} & 63.1 & 71.2 & 75.5 & 70.4 & \underline{75.8} & \underline{76.9}  \\
  ~~DELG~\cite{CaoX20UnifyingDeepLocalGlobalFeatures}+SIFT~\cite{Lowe04IJCV}          & 79.3 & 84.4 & 86.2 & 90.0 & 91.5 & 92.5 & 97.8 & 99.9 & \textbf{100.0} & 61.6 & 69.2 & 73.8 & 69.0 & 74.6 & 75.6  \\
  ~~DenseVLAD~\cite{ToriiCVPR15PlaceRecognitionByViewSynthesis}+SIFT~\cite{Lowe04IJCV}     & 85.5 & 90.2 & 91.2 & 91.3 & 92.5 & 93.0 & 97.8 & 99.9 & 99.9 & 62.4 & 70.0 & 73.6 & 64.8 & 69.8 & 70.6  \\
  ~~NetVLAD~\cite{ArandjelovicCVPR16NetVLADPlaceRecognition}+SIFT~\cite{Lowe04IJCV}       & 83.2 & 88.4 & 90.0 & 91.0 & 92.2 & 92.9 & 97.7 & \textbf{100.0} & \textbf{100.0} & 58.8 & 66.5 & 71.0 & 67.2 & 72.4 & 73.4  \\
  ~~SIFT~\cite{Lowe04IJCV}+vocab.~tree (COLMAP~\cite{schoenberger2016sfm}) & 69.8 & 77.7 & 80.3 & 89.5 & 91.0 & 92.2 & 97.6 & 99.8 & 99.8 & 56.2 & 65.6 & 70.3 & 54.9 & 59.2 & 60.6  \\
  ~~APGeM~\cite{RevaudICCV19LearningwithAPTrainingImgRetrievalListwiseLoss}+D2-Net~\cite{DusmanuCVPR19D2NetDeepLocalFeatures}           & 87.0 & 91.8 & 92.9 & \underline{95.8} & \underline{96.6} & \underline{97.0} & 97.5 & \textbf{100.0} & \textbf{100.0} & 68.0 & 77.8 & \underline{81.4} & 68.7 & 75.2 & 76.5  \\
  ~~DELG~\cite{CaoX20UnifyingDeepLocalGlobalFeatures}+D2-Net~\cite{DusmanuCVPR19D2NetDeepLocalFeatures}            & 85.7 & 89.6 & 90.4 & 94.2 & 95.4 & 95.7 & 97.6 & 99.7 & 99.9 & 67.2 & 76.7 & 80.0 & 67.5 & 73.0 & 74.2  \\
  ~~DenseVLAD~\cite{ToriiCVPR15PlaceRecognitionByViewSynthesis}+D2-Net~\cite{DusmanuCVPR19D2NetDeepLocalFeatures}       & \underline{91.5} & \underline{94.3} & \underline{94.9} & 92.9 & 93.9 & 94.1 & 97.3 & 99.3 & 99.3 & 68.4 & 76.0 & 79.2 & 61.9 & 67.2 & 68.5  \\
  ~~NetVLAD~\cite{ArandjelovicCVPR16NetVLADPlaceRecognition}+D2-Net~\cite{DusmanuCVPR19D2NetDeepLocalFeatures}         & 89.7 & 93.1 & 94.1 & 93.9 & 95.2 & 95.7 & 97.5 & 99.8 & 99.9 & 64.0 & 72.4 & 75.9 & 65.3 & 70.8 & 71.5  \\
  ~~APGeM~\cite{RevaudICCV19LearningwithAPTrainingImgRetrievalListwiseLoss}+R2D2~\cite{RevaudNIPS19R2D2ReliableRepeatableDetectorsDescriptors}         & 89.6 & 92.6 & 93.4 & \textbf{96.4} & \textbf{97.2} & \textbf{97.4} & \textbf{98.0} & \textbf{100.0} & \textbf{100.0} & \underline{70.0} & \textbf{78.7} & \textbf{82.5} & \textbf{72.7} & \textbf{77.9} & \textbf{78.6}  \\
  ~~DELG~\cite{CaoX20UnifyingDeepLocalGlobalFeatures}+R2D2~\cite{RevaudNIPS19R2D2ReliableRepeatableDetectorsDescriptors}          & 87.1 & 90.2 & 91.2 & 95.7 & 96.1 & 96.5 & \textbf{98.0} & \textbf{100.0} & \textbf{100.0} & 68.4 & 77.0 & 80.5 & \underline{70.7} & 75.7 & 76.2  \\
  ~~DenseVLAD~\cite{ToriiCVPR15PlaceRecognitionByViewSynthesis}+R2D2~\cite{RevaudNIPS19R2D2ReliableRepeatableDetectorsDescriptors}     & \textbf{92.2} & \textbf{95.1} & \textbf{95.7} & 95.4 & 96.2 & 96.3 & 97.8 & 99.9 & 99.9 & \textbf{70.2} & \underline{78.0} & 81.0 & 65.5 & 69.8 & 70.5  \\
  ~~NetVLAD~\cite{ArandjelovicCVPR16NetVLADPlaceRecognition}+R2D2~\cite{RevaudNIPS19R2D2ReliableRepeatableDetectorsDescriptors}       & 91.1 & 93.9 & 94.8 & 95.6 & 96.4 & \underline{97.0} & 97.8 & \textbf{100.0} & \textbf{100.0} & 65.8 & 73.6 & 77.2 & 68.4 & 72.8 & 73.3  \\
\hline
\textit{ESAC~\cite{Brachmann2019ICCVa}} &&&&&&&&&&&&&&& \\
  ~~1 expert           & 0.0 & 0.0 & 0.1 & 0.0 & 0.1 & 2.2 & 1.2 & 12.5 & 55.0 & 0.0 & 0.0 & 0.0 & 0.0 & 0.0 & 0.0  \\
  ~~10 experts         & 0.0 & 0.6 & 4.2 & 19.4 & 36.4 & 52.4 & 86.9 & 98.2 & 99.8 & 10.8 & 30.1 & 49.0 & 24.3 & 41.2 & 55.4  \\
  ~~20 experts         & 0.8 & 3.5 & 10.4 & 32.7 & 48.5 & 61.2 & 89.4 & 98.5 & 99.8 & 26.1 & 48.1 & 63.7 & 27.6 & 45.7 & 55.8  \\
  ~~50 experts         & 5.4 & 13.3 & 23.7 & 44.8 & 60.5 & 69.9 & 89.7 & 99.2 & 99.9 & 35.8 & 56.7 & 68.8 & 26.7 & 42.5 & 52.5  \\
\hline
  \textit{PoseNet~\cite{KendallICCV15PoseNetCameraRelocalization}}     & 0.0 & 0.0 & 0.1 & 0.0 & 0.0 & 0.4 & 0.0 & 0.0 & 0.5 & 0.0 & 0.0 & 0.2 & 0.0 & 0.0 & 0.1  \\
\hline
\end{tabular}
}
\caption{Results of various visual localization methods on the 5 NAVER LABS datasets, with the percentages of successfully localized test images within three thresholds for each datasets. The best method is shown in bold, the second best is underlined. We report the results for the Basler images of the test set (top), the Galaxy images of the validation set (middle) and the Basler images of the validation (bottom).}
\label{tab:big}
\end{table*}

\begin{figure*}
\setlength{\tabcolsep}{2pt}
\resizebox{\linewidth}{!}{
\begin{tabular}{cccccc}
  & Dept. B1 & Dept. 1F & Dept. 4F & Metro St. B1 & Metro St. B2 \\
 1st & \includegraphics[width=0.25\paperwidth]{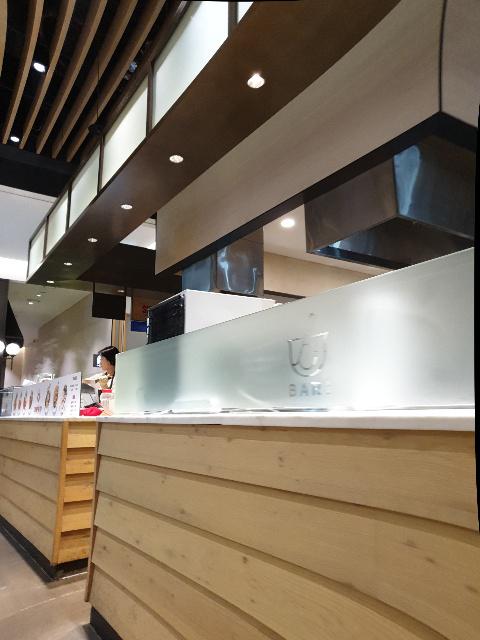}  & \includegraphics[width=0.25\paperwidth]{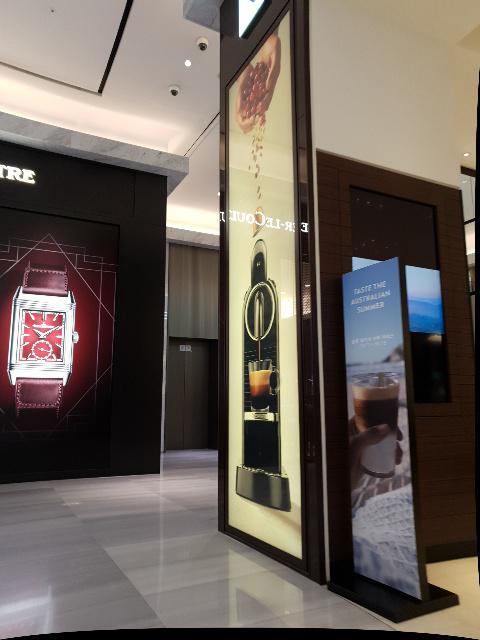}  & \includegraphics[width=0.25\paperwidth]{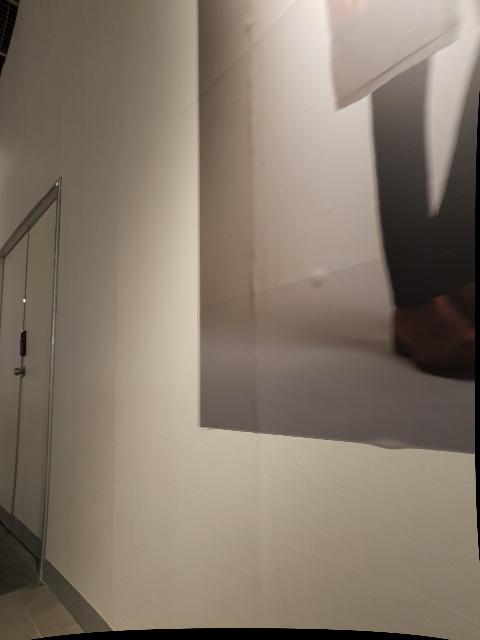}  & \includegraphics[width=0.25\paperwidth]{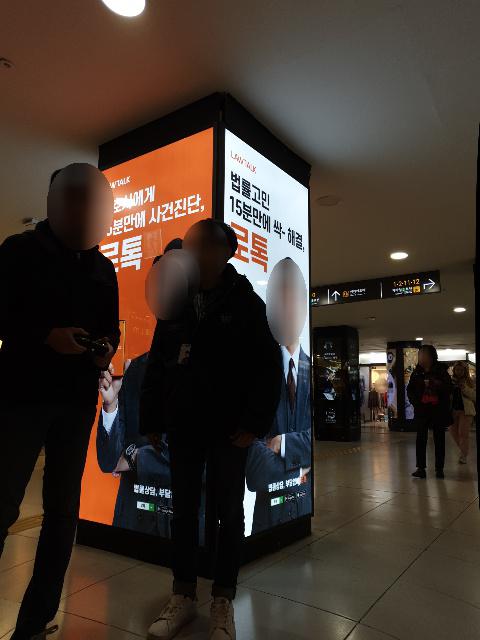}  & \includegraphics[width=0.25\paperwidth]{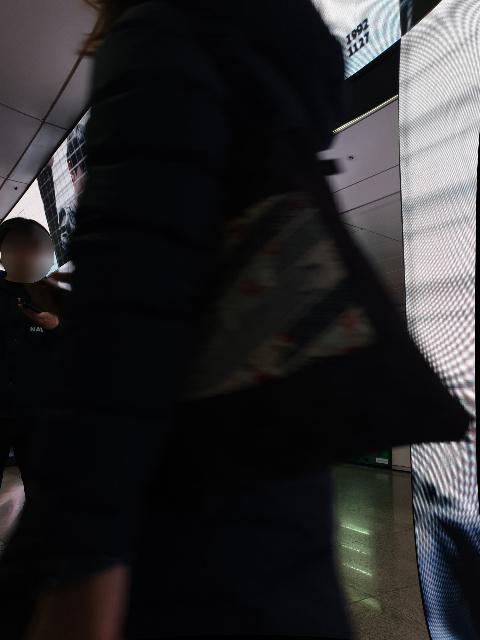} \\
  & pos. error: 22893.834m  & pos. error: 9423.427m  & pos. error: 2320.612m  & pos. error: 56576.396m  & pos. error: 2522.412m \\
  & \textcolor{green}{low freq. score: 25.94}  & \textcolor{green}{low freq. score: 27.04}  & \textcolor{red}{low freq. score: 2.36}  & \textcolor{green}{low freq. score: 20.19}  & \textcolor{red}{low freq. score: 9.61} \\
  & \textcolor{green}{crowdedness: 0.00}  & \textcolor{green}{crowdedness: 0.00}  & \textcolor{green}{crowdedness: 0.00}  & \textcolor{red}{crowdedness: 25.05}  & \textcolor{red}{crowdedness: 56.35} \\
 2nd & \includegraphics[width=0.25\paperwidth]{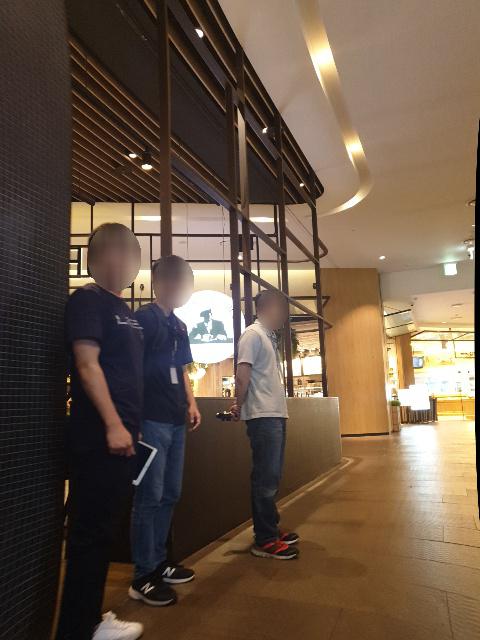}  & \includegraphics[width=0.25\paperwidth]{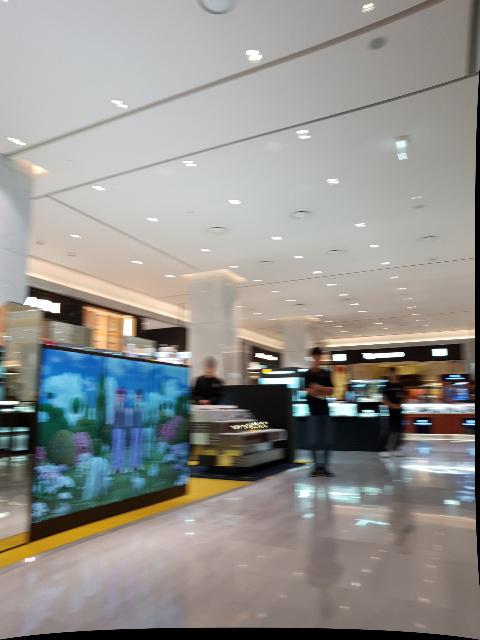}  & \includegraphics[width=0.25\paperwidth]{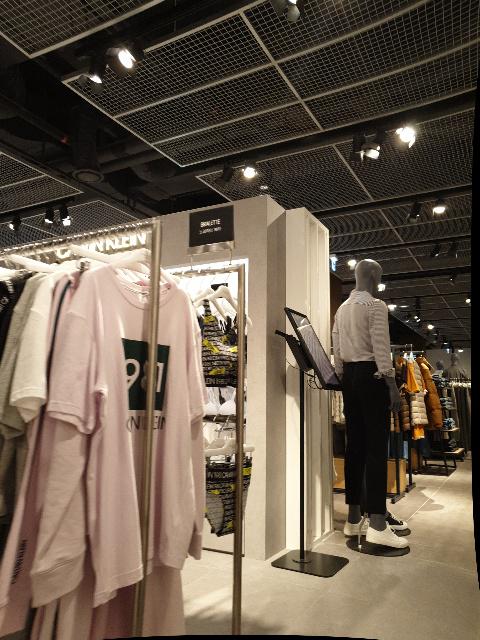}  & \includegraphics[width=0.25\paperwidth]{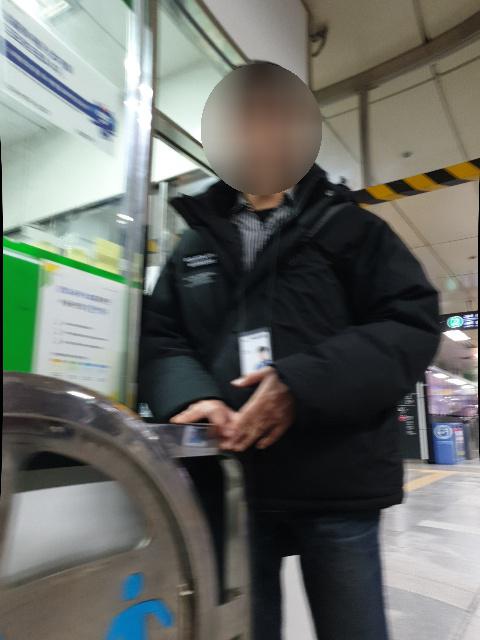}  & \includegraphics[width=0.25\paperwidth]{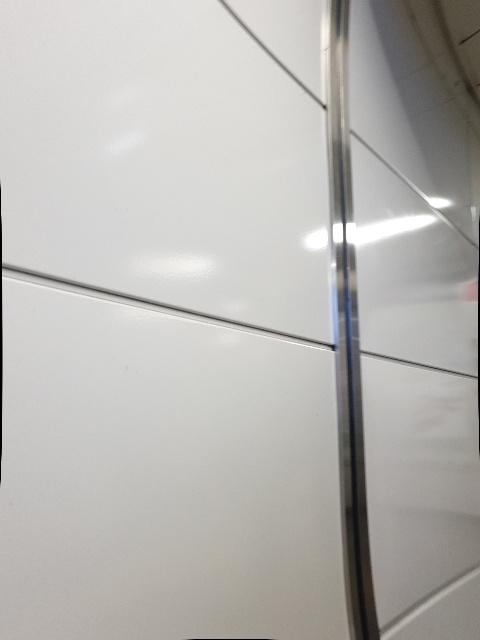} \\
  & pos. error: 17471.276m  & pos. error: 3469.674m  & pos. error: 1227.543m  & pos. error: 1730.929m  & pos. error: 831.310m \\
  & \textcolor{green}{low freq. score: 27.29}  & \textcolor{green}{low freq. score: 21.97}  & \textcolor{green}{low freq. score: 41.27}  & \textcolor{green}{low freq. score: 20.98}  & \textcolor{red}{low freq. score: 9.48} \\
  & \textcolor{green}{crowdedness: 16.95}  & \textcolor{green}{crowdedness: 0.14}  & \textcolor{green}{crowdedness: 0.94}  & \textcolor{red}{crowdedness: 38.38}  & \textcolor{green}{crowdedness: 0.00} \\
 3rd & \includegraphics[width=0.25\paperwidth]{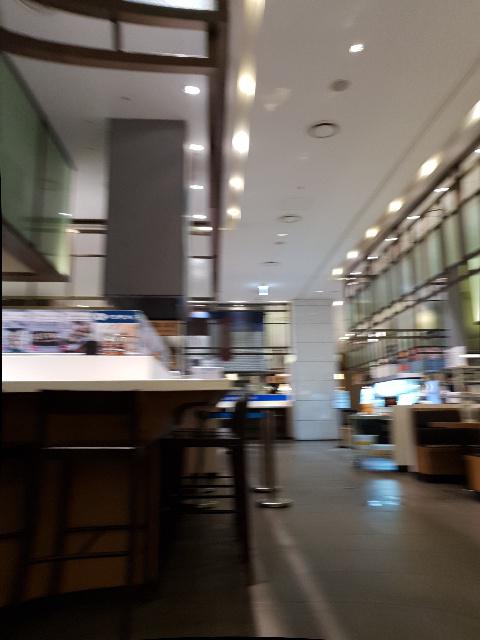}  & \includegraphics[width=0.25\paperwidth]{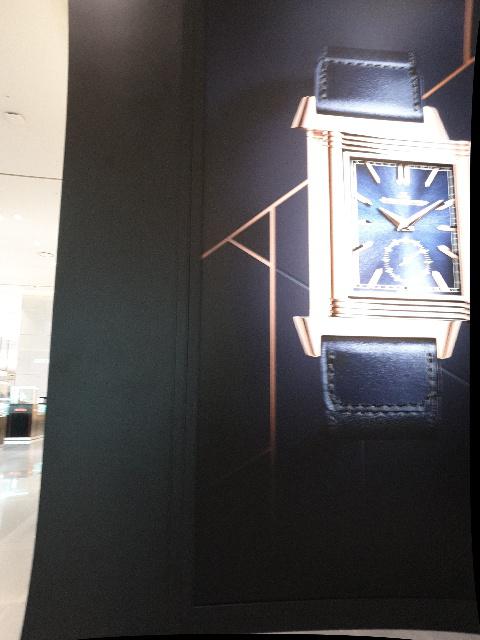}  & \includegraphics[width=0.25\paperwidth]{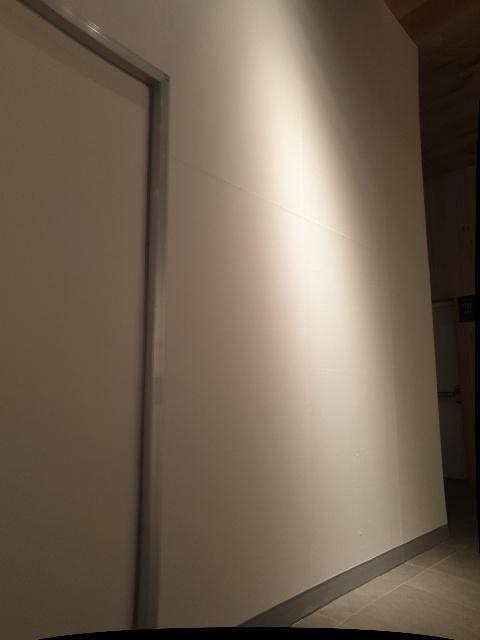}  & \includegraphics[width=0.25\paperwidth]{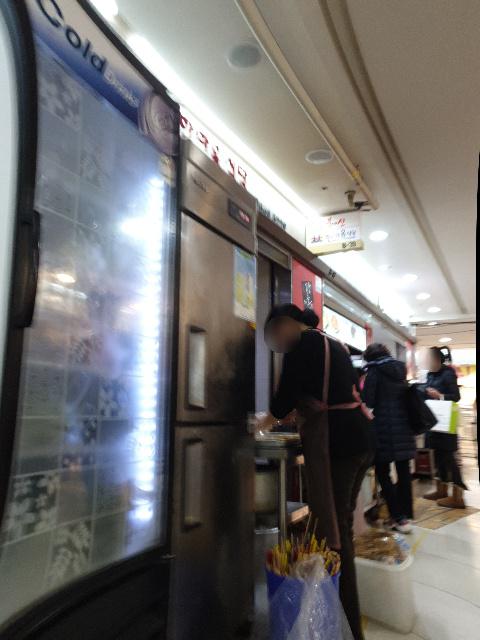}  & \includegraphics[width=0.25\paperwidth]{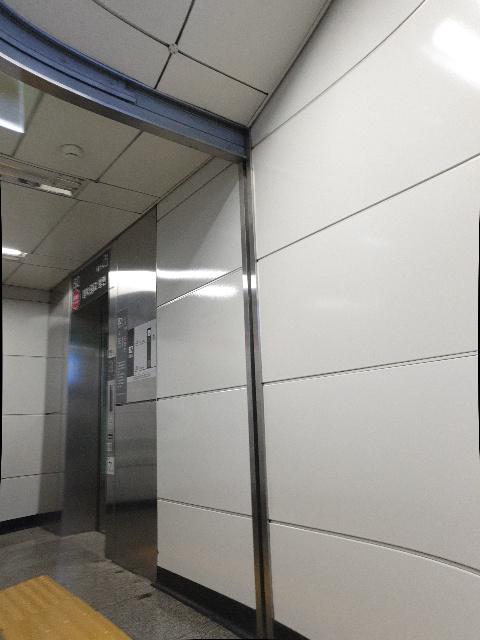} \\
  & pos. error: 872.742m  & pos. error: 743.336m  & pos. error: 684.469m  & pos. error: 1598.624m  & pos. error: 587.146m \\
  & \textcolor{green}{low freq. score: 23.12}  & \textcolor{green}{low freq. score: 27.03}  & \textcolor{red}{low freq. score: 0.41}  & \textcolor{green}{low freq. score: 23.90}  & \textcolor{green}{low freq. score: 22.47} \\
  & \textcolor{green}{crowdedness: 0.00}  & \textcolor{green}{crowdedness: 0.00}  & \textcolor{green}{crowdedness: 0.00}  & \textcolor{green}{crowdedness: 5.54}  & \textcolor{green}{crowdedness: 0.00} \\
 4th & \includegraphics[width=0.25\paperwidth]{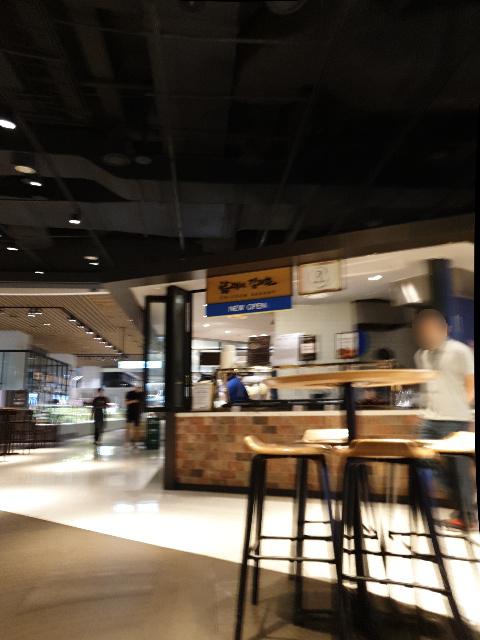}  & \includegraphics[width=0.25\paperwidth]{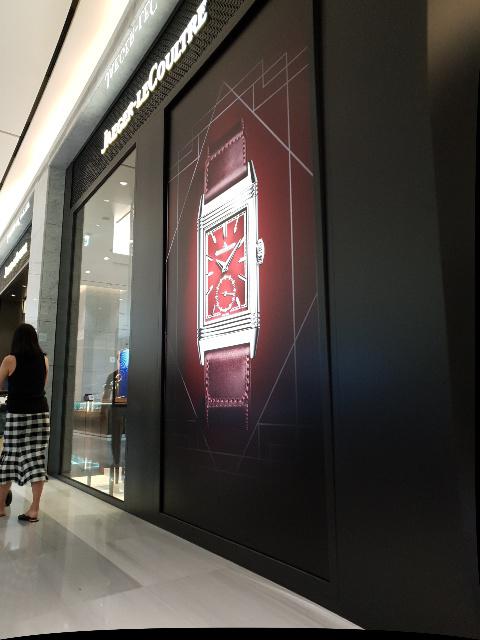}  & \includegraphics[width=0.25\paperwidth]{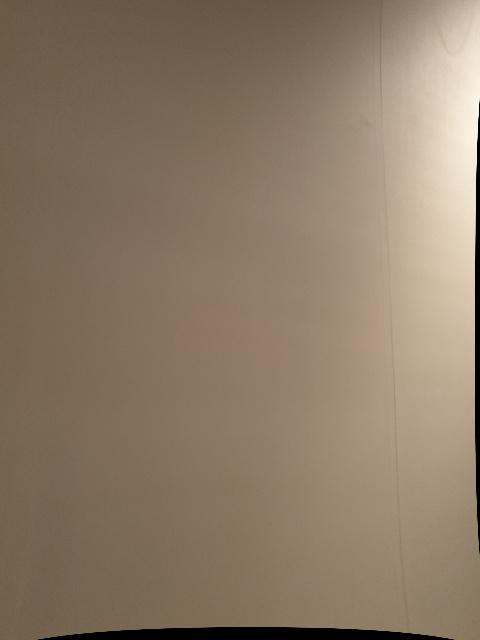}  & \includegraphics[width=0.25\paperwidth]{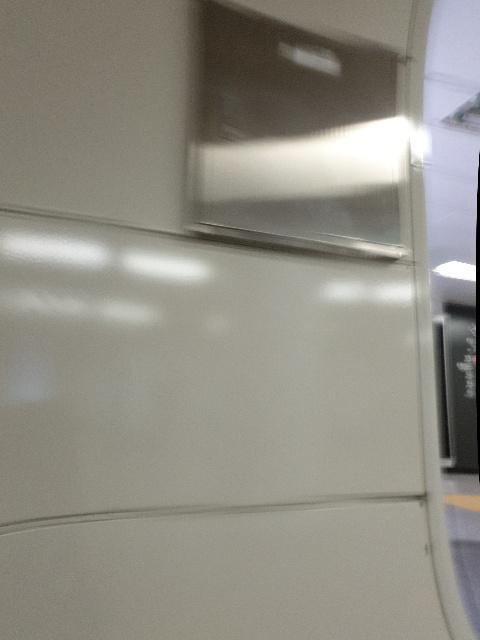}  & \includegraphics[width=0.25\paperwidth]{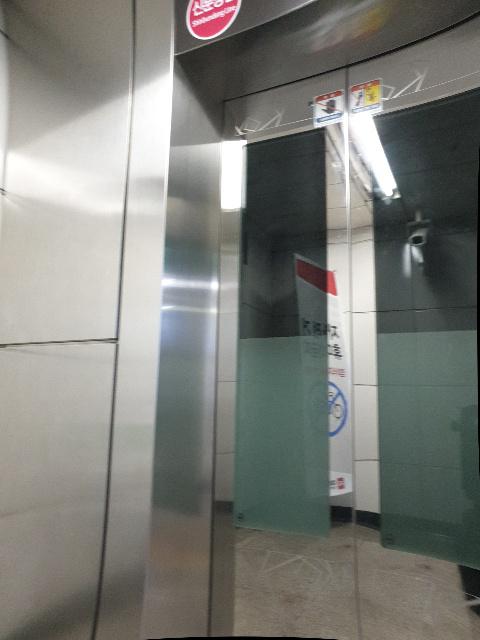} \\
  & pos. error: 869.867m  & pos. error: 540.344m  & pos. error: 588.486m  & pos. error: 1442.370m  & pos. error: 421.718m \\
  & \textcolor{green}{low freq. score: 28.01}  & \textcolor{green}{low freq. score: 23.50}  & \textcolor{red}{low freq. score: -3.48}  & \textcolor{red}{low freq. score: 12.67}  & \textcolor{green}{low freq. score: 21.13} \\
  & \textcolor{green}{crowdedness: 2.59}  & \textcolor{green}{crowdedness: 1.77}  & \textcolor{green}{crowdedness: 0.00}  & \textcolor{green}{crowdedness: 0.78}  & \textcolor{green}{crowdedness: 0.00} \\a
\end{tabular}
}
\caption{Worst localized images according to the positional error for DELG+R2D2 on the 5 NAVER LABS localization datasets (test set - Galaxy images). Red: low freq.~score below 20, crowdedness above 20}
\label{fig:worst}
\end{figure*}

\begin{figure*}
\setlength{\tabcolsep}{2pt}
\resizebox{\linewidth}{!}{
\begin{tabular}{cccccc}
  & Dept. B1 & Dept. 1F & Dept. 4F & Metro St. B1 & Metro St. B2 \\
 1st & \includegraphics[width=0.25\paperwidth]{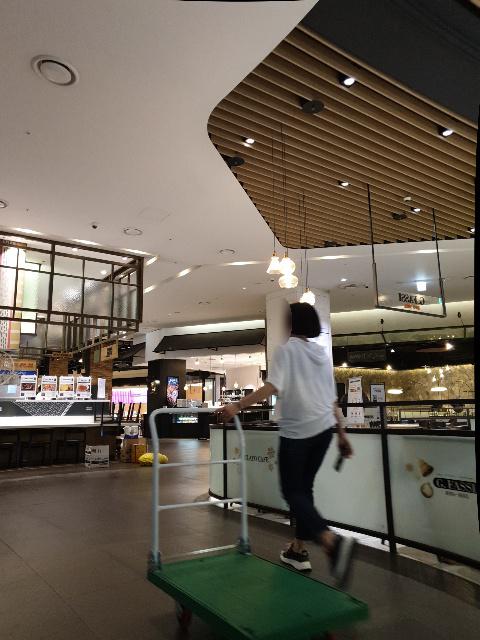}  & \includegraphics[width=0.25\paperwidth]{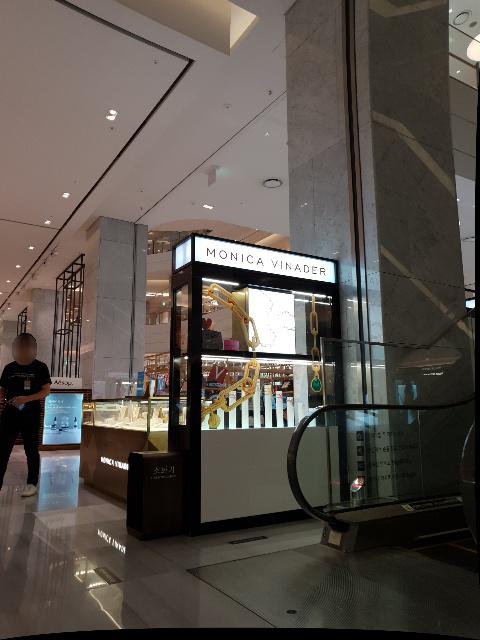}  & \includegraphics[width=0.25\paperwidth]{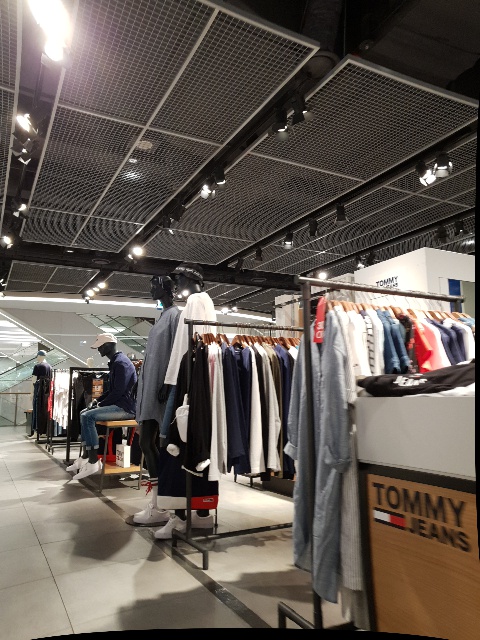}  & \includegraphics[width=0.25\paperwidth]{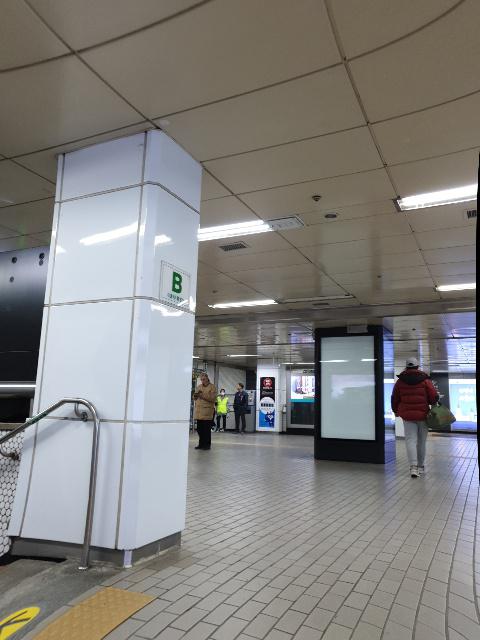}  & \includegraphics[width=0.25\paperwidth]{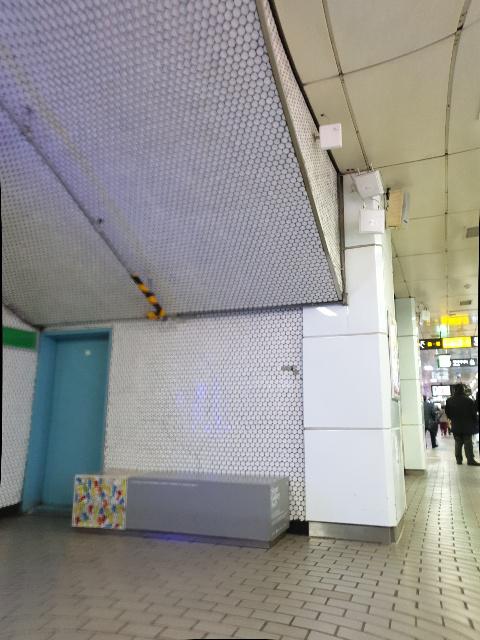} \\
  & pos. error: 0.002m  & pos. error: 0.001m  & pos. error: 0.000m  & pos. error: 0.002m  & pos. error: 0.004m \\
  & \textcolor{green}{low freq. score: 34.43}  & \textcolor{green}{low freq. score: 27.25}  & \textcolor{green}{low freq. score: 42.22}  & \textcolor{green}{low freq. score: 25.95}  & \textcolor{green}{low freq. score: 29.67} \\
  & \textcolor{green}{crowdedness: 5.11}  & \textcolor{green}{crowdedness: 1.72}  & \textcolor{green}{crowdedness: 1.63}  & \textcolor{green}{crowdedness: 0.00}  & \textcolor{green}{crowdedness: 0.73} \\
 2nd & \includegraphics[width=0.25\paperwidth]{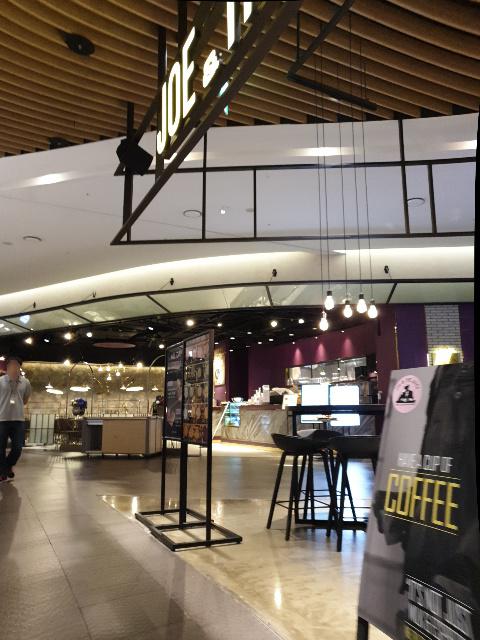}  & \includegraphics[width=0.25\paperwidth]{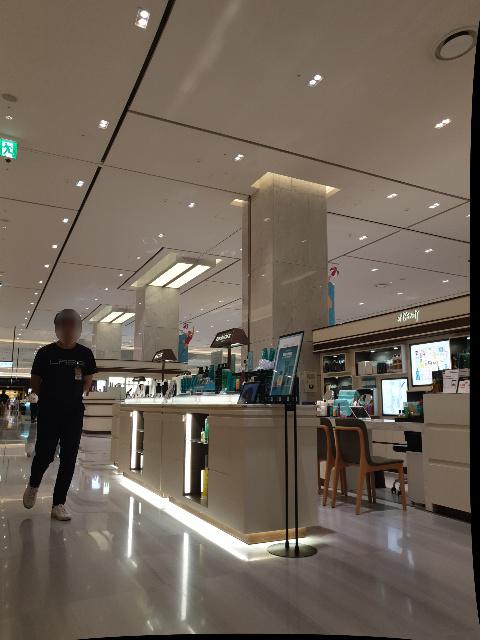}  & \includegraphics[width=0.25\paperwidth]{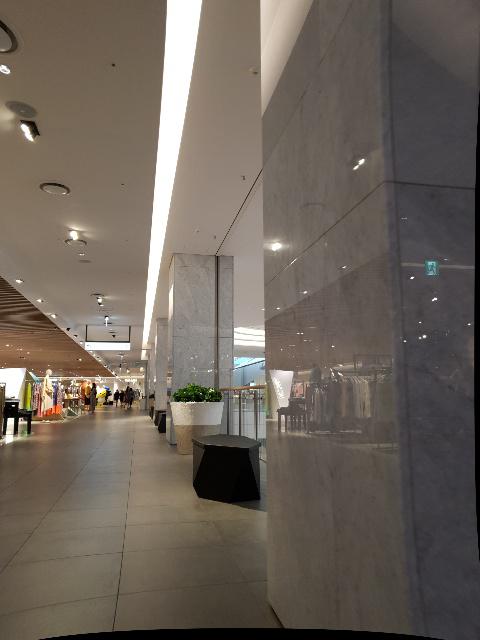}  & \includegraphics[width=0.25\paperwidth]{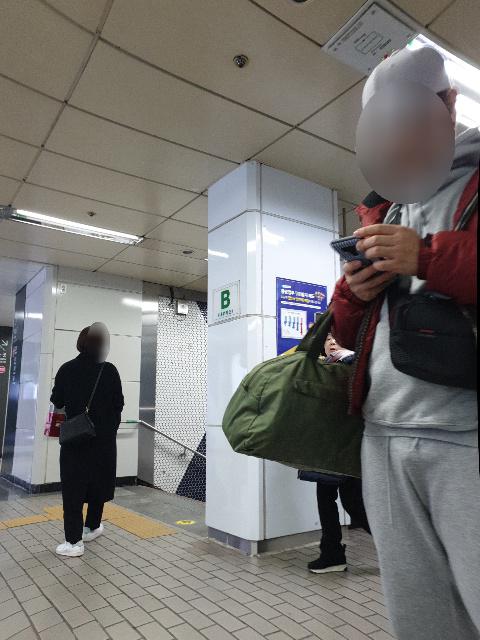}  & \includegraphics[width=0.25\paperwidth]{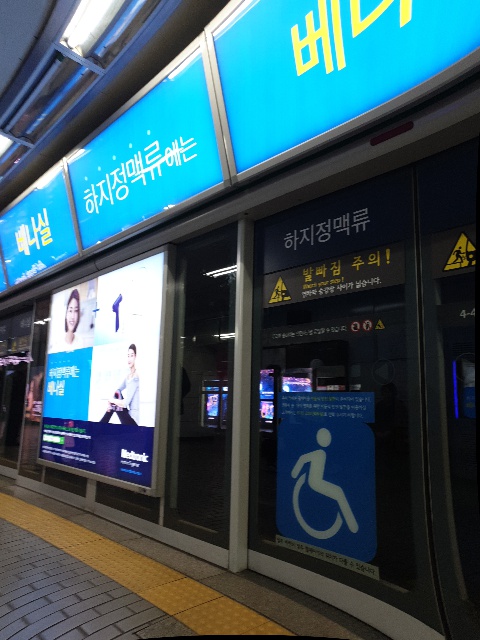} \\
  & pos. error: 0.003m  & pos. error: 0.001m  & pos. error: 0.000m  & pos. error: 0.003m  & pos. error: 0.004m \\
  & \textcolor{green}{low freq. score: 39.40}  & \textcolor{green}{low freq. score: 24.77}  & \textcolor{red}{low freq. score: 15.87}  & \textcolor{green}{low freq. score: 31.33}  & \textcolor{green}{low freq. score: 28.38} \\
  & \textcolor{green}{crowdedness: 0.73}  & \textcolor{green}{crowdedness: 0.59}  & \textcolor{green}{crowdedness: 0.00}  & \textcolor{red}{crowdedness: 23.67}  & \textcolor{green}{crowdedness: 0.47} \\
 3rd & \includegraphics[width=0.25\paperwidth]{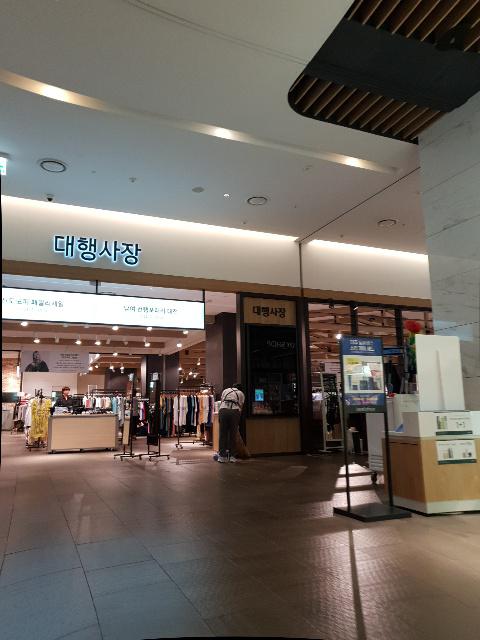}  & \includegraphics[width=0.25\paperwidth]{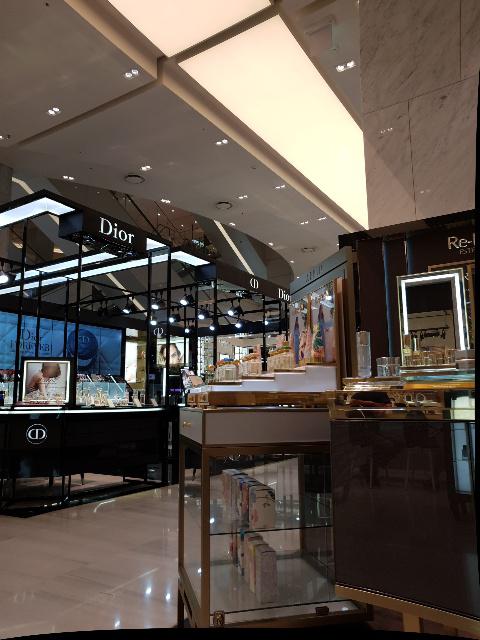}  & \includegraphics[width=0.25\paperwidth]{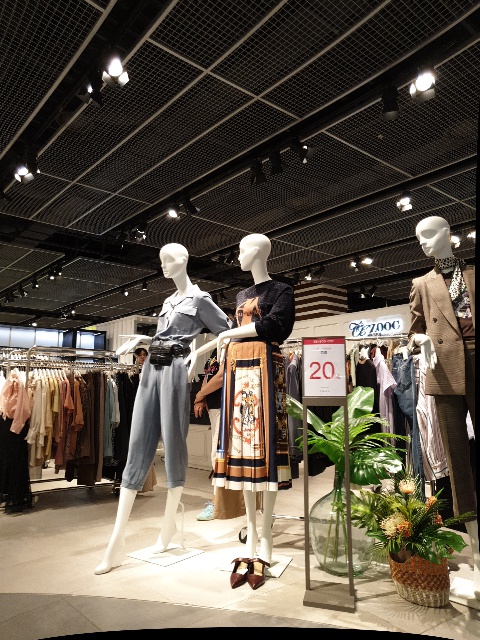}  & \includegraphics[width=0.25\paperwidth]{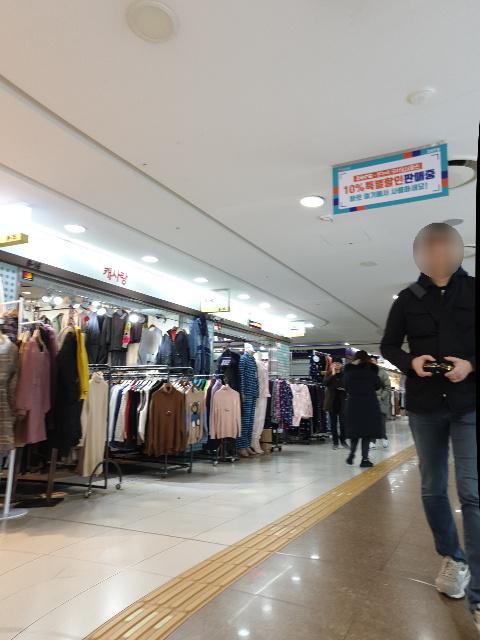}  & \includegraphics[width=0.25\paperwidth]{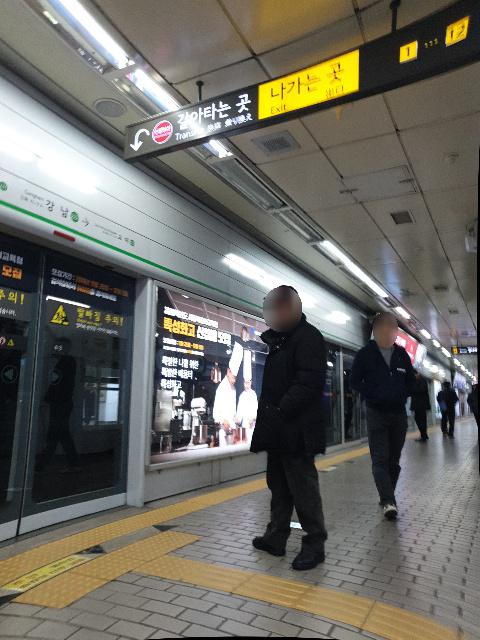} \\
  & pos. error: 0.003m  & pos. error: 0.001m  & pos. error: 0.000m  & pos. error: 0.004m  & pos. error: 0.005m \\
  & \textcolor{green}{low freq. score: 27.13}  & \textcolor{green}{low freq. score: 32.07}  & \textcolor{green}{low freq. score: 47.13}  & \textcolor{green}{low freq. score: 23.87}  & \textcolor{green}{low freq. score: 33.87} \\
  & \textcolor{green}{crowdedness: 1.09}  & \textcolor{green}{crowdedness: 0.27}  & \textcolor{green}{crowdedness: 4.60}  & \textcolor{green}{crowdedness: 5.97}  & \textcolor{green}{crowdedness: 8.40} \\
 4th & \includegraphics[width=0.25\paperwidth]{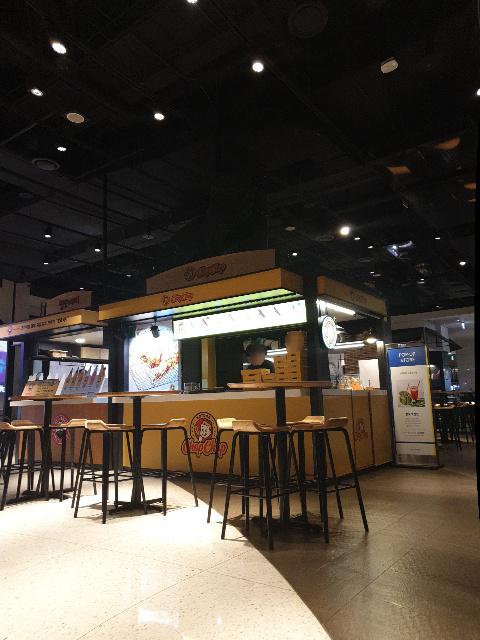}  & \includegraphics[width=0.25\paperwidth]{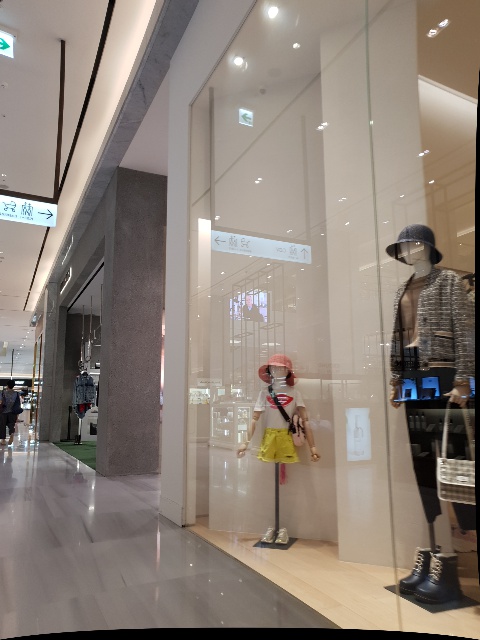}  & \includegraphics[width=0.25\paperwidth]{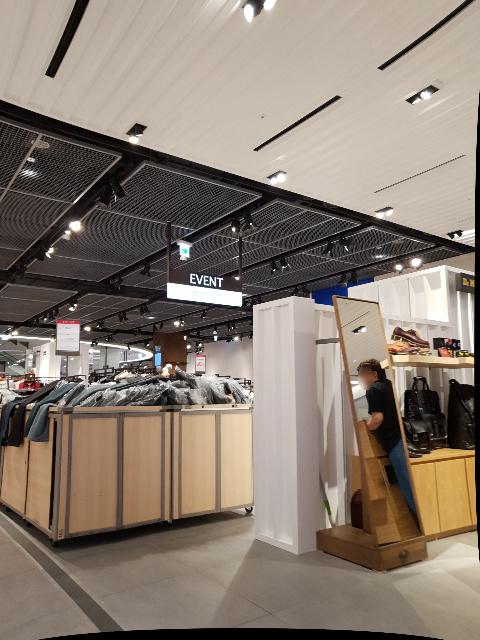}  & \includegraphics[width=0.25\paperwidth]{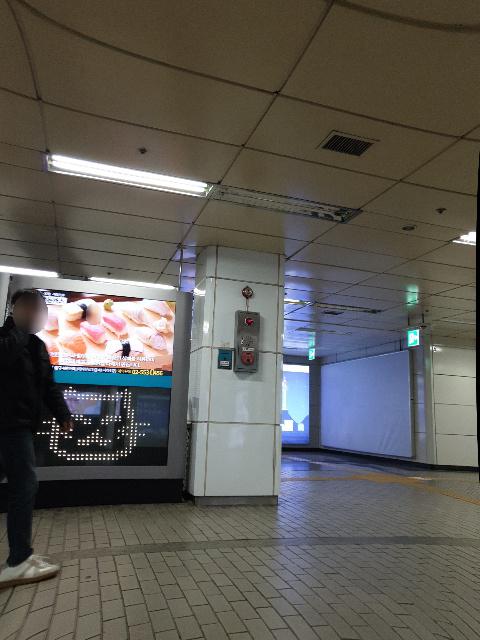}  & \includegraphics[width=0.25\paperwidth]{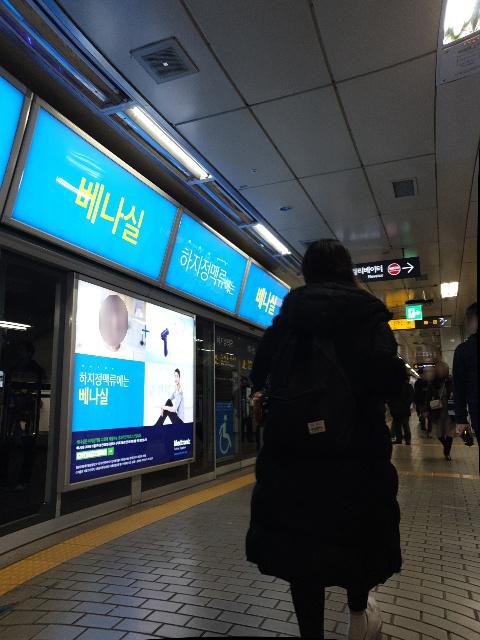} \\
  & pos. error: 0.003m  & pos. error: 0.001m  & pos. error: 0.000m  & pos. error: 0.004m  & pos. error: 0.006m \\
  & \textcolor{green}{low freq. score: 30.90}  & \textcolor{green}{low freq. score: 22.06}  & \textcolor{green}{low freq. score: 34.61}  & \textcolor{green}{low freq. score: 27.83}  & \textcolor{green}{low freq. score: 28.25} \\
  & \textcolor{green}{crowdedness: 0.00}  & \textcolor{green}{crowdedness: 0.88}  & \textcolor{green}{crowdedness: 1.68}  & \textcolor{green}{crowdedness: 3.30}  & \textcolor{green}{crowdedness: 11.85} \\
\end{tabular}

}
\caption{Best localized images according to the positional error for DELG+R2D2 on the 5 NAVER LABS localization datasets (test set - Galaxy images). Red: low freq.~score below 20, crowdedness above 20}
\label{fig:best}
\end{figure*}

\noindent \textbf{Qualitative results.}
To further analyze the results of DELG+R2D2, one of the best methods according to our experiments, in Figure~\ref{fig:worst} (resp.~Figure~\ref{fig:best}) we show the 4 images with the highest (resp.~lowest) positional error for each of the 5 NAVER LABS localization datasets.
We observe that many images that are not localized either lack relevant information, especially in Dept.~4F (which can be often seen in the low freq.~score of these images), contain people occluding the images (which can be seen in the crowdedness score), or contain large changing elements (\eg large screen with varying content).
In contrast, images that are well localized contain lot of high frequency and relevant information.
They also contain some dynamic elements but these are not dominant in the images.
Note that these observations cannot be generalized because the localization performance also depends on the content of the images.
An image with little low frequency content, for example, can still be localized precisely if the combination of visual information can be uniquely described in the dataset and robustly recovered during the localization process.

\clearpage
{\small
\bibliographystyle{ieee_fullname}
\bibliography{references}
}

\end{document}